\documentclass{article}

\PassOptionsToPackage{numbers,sort&compress}{natbib}
\PassOptionsToPackage{hidelinks}{hyperref}
\usepackage[final]{neurips_data_2021}

\usepackage{amsmath} 
\usepackage{graphicx}
\usepackage{wrapfig}
\usepackage{subcaption}

\usepackage[utf8]{inputenc} 
\usepackage[T1]{fontenc}    
\usepackage{hyperref}       
\usepackage{url}            
\usepackage{booktabs}       
\usepackage{amsfonts}       
\usepackage{nicefrac}       
\usepackage{microtype}      
\usepackage{xcolor}         

\newcommand{\abs}[1]{\left| #1 \right|} 
 
\title{Chaos as an interpretable benchmark for forecasting and data-driven modelling}
\author{%
  William Gilpin\thanks{Dataset and benchmark available at: https://github.com/williamgilpin/dysts} \\
  Department of Physics \& Oden Institute, UT Austin\\
  Quantitative Biology Initiative, Harvard University\\
  \texttt{wgilpin@fas.harvard.edu} \\
}

\begin{document}

\maketitle

\begin{abstract}


The striking fractal geometry of strange attractors underscores the generative nature of chaos: like probability distributions, chaotic systems can be repeatedly measured to produce arbitrarily-detailed information about the underlying attractor. Chaotic systems thus pose a unique challenge to modern statistical learning techniques, while retaining quantifiable mathematical properties that make them controllable and interpretable as benchmarks. Here, we present a growing database currently comprising 131 known chaotic dynamical systems spanning fields such as astrophysics, climatology, and biochemistry. Each system is paired with precomputed multivariate and univariate time series. Our dataset has comparable scale to existing static time series databases; however, our systems can be re-integrated to produce additional datasets of arbitrary length and granularity. Our dataset is annotated with known mathematical properties of each system, and we perform feature analysis to broadly categorize the diverse dynamics present across the collection. Chaotic systems inherently challenge forecasting models, and across extensive benchmarks we correlate forecasting performance with the degree of chaos present.  We also exploit the unique generative properties of our dataset in several proof-of-concept experiments: surrogate transfer learning to improve time series classification, importance sampling to accelerate model training, and benchmarking symbolic regression algorithms.

\end{abstract}

\section{Introduction}

Two trajectories emanating from distinct locations on a strange attractor will never recur nor intersect, a basic mathematical property that underlies the complex geometry of chaos. As a result, measurements drawn from a chaotic system are deterministic yet non-repeating, even at finite resolution \cite{crutchfield1982symbolic,cvitanovic2005chaos}. Thus, while representations of chaotic systems are finite (e.g. differential equations or discrete maps), they can indefinitely generate new data, allowing the fractal structure of the attractor to be resolved in ever-increasing detail \cite{farmer1982information}. This interplay between the boundedness of the attractor and non-recurrence of the dynamics is responsible for the complexity of diverse systems, ranging from the intricate gyrations of orbiting stars to the irregular spiking of neuronal ensembles \cite{grebogi1987chaos,ott2002chaos}.

Chaotic systems thus represent a unique testbed for modern statistical learning techniques. Their unpredictability challenges traditional forecasting methods, while their fractal geometry precludes concise representations \cite{tang2020introduction}. While modeling and forecasting chaos remains a fundamental problem in its own right \cite{pathak2018model,boffetta2002predictability}, many prior works on general time series analysis and data-driven model inference have used specific chaotic systems (such as the Lorenz "butterfly" attractor) as toy problems in order to demonstrate method performance in a controlled setting \cite{nassar2018tree,champion2019data,costa2019adaptive,gilpin2020deep,greydanus2019hamiltonian,lu2020supervised,yu2017long,lu2017reservoir,bellot2021consistency,wang2021reconstructing,li2020scalable,ma2007chaotic}. In this context, there are several advantages to chaotic systems as benchmarks for time series analysis and data-driven modelling: \\
(1) Chaotic systems have provably complex dynamics, which arise due to underlying mathematical structure, rather than abstruse representations of otherwise simple latent dynamics. \\
(2) Existing time series databases contain datasets chosen primarily for availability or applicability, rather than for having innate properties (e.g. complexity, quasiperiodicity, dimensionality) that span the range of possible behaviors time series may exhibit. \\
(3) Chaotic systems have accessible generating processes, making it possible to obtain new data and representations, and for the benchmark to be related to mechanistic details of the underlying system. \\
These properties suggest that chaotic systems can aid in interpreting the properties of complex models \cite{tang2020introduction,kantz2004nonlinear}. However, chaotic systems as benchmarks lack standardization, and prior works' emphasis on single systems like the Lorenz attractor may undermine generalizability. Moreover, focusing on isolated systems neglects the diversity of dynamics in known chaotic systems, thereby preventing systematic quantification and interpretation of algorithm performance relative to the mathematical properties of different systems.

Here, we present a growing database of low-dimensional chaotic systems drawn from published work in diverse domains such as meteorology, neuroscience, hydrodynamics, and astrophysics. Each system is represented by several multivariate time series drawn from the dynamics, annotations of known mathematical properties, and an explicit analytical form that can be re-integrated to generate new time series of arbitrary length, stochasticity, and granularity. We provide extensive forecasting benchmarks across our systems, allowing us to interpret the empirical performance of different forecasting techniques in the context of mathematical properties such as system chaoticity. Our dataset improves the interpretability of time series algorithms by allowing methods to be compared across time series with different intrinsic properties and underlying generating processes---thereby complementing existing interpretability methods that identify salient feature sets or time windows within single time series \cite{ismail2020benchmarking,lim2021temporal}. We also consider applications to data-driven modelling in the form of symbolic regression and neural ordinary differential equations tasks, and we show the surprising result that the accuracy of a symbolic regression-derived formula can correlate with mathematical properties of the dynamics produced by the formula. Finally, we demonstrate unique applications enabled by the ability to re-integrate our dataset: we pre-train a timescale-matched feature extractor for an existing time series classification benchmark, and we accelerate training of a forecast model by importance sampling sparse regions on the dynamical attractor.

\section{Description of Datasets}

\begin{figure}
  \centering
 \includegraphics[width=\linewidth]{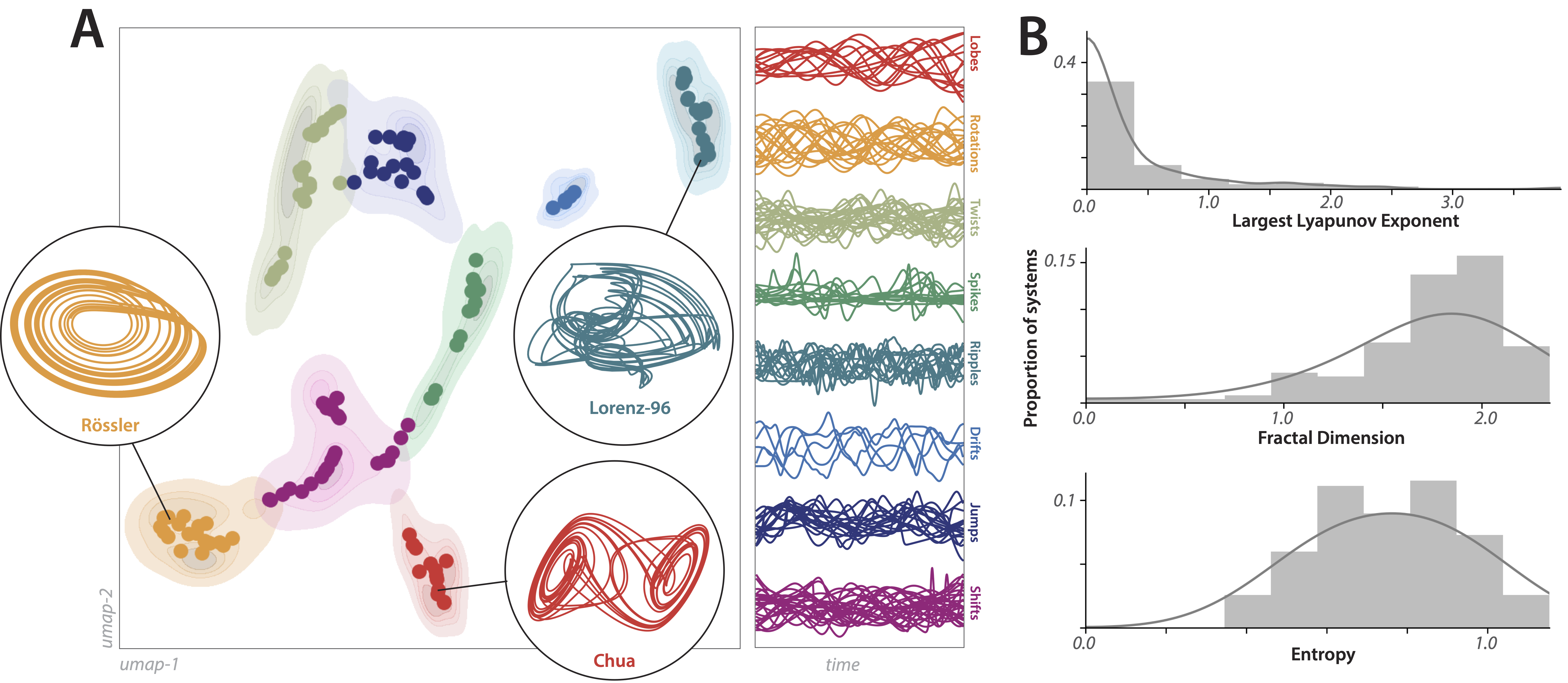}
  \caption{{\bf Properties of the chaotic dynamical systems dataset.} (A) Embeddings of $131$ chaotic dynamical systems. Points correspond to average embeddings of individual systems, and shading shows ranges over many random initial conditions. Colors correspond to an unsupervised clustering, and example dynamics for each cluster are shown. (B) Distributions of key mathematical properties across the dataset.
 }
 \label{fig_stats}
 \vspace{-6mm}
\end{figure}

\paragraph{Scope.} The diverse dynamical systems in our dataset span astrophysics, neuroscience, ecology, climatology, hydrodynamics, and many other domains. The supplementary material contains a glossary defining key terms from dynamical systems theory relevant to our dataset. Each entry in our dataset represents a single dynamical system, such as the Lorenz attractor, that takes an initial condition as input and outputs a trajectory representing the input point's location as time evolves. Systems are chosen based on prior appearance as named systems in published works. In order to provide a consistent test for time series models, we define chaos in the mathematical sense: two copies of a system prepared in infinitesimally different initial states will exponentially diverge over time. We also focus particularly on chaotic systems that produce low-dimensional strange attractors, which are fractal structures that display bounded, stationary, and ergodic dynamics with quantifiable mathematical properties. As a result, we exclude transient chaos and chaotic {\it repellers} (chaotic regions that trajectories eventually escape) \cite{tel2015joy,chen2017slim,grebogi1986critical}, as well as most nonchaotic strange attractors save for one paradigmatic example: a quasiperiodic two-dimensional torus \cite{grebogi1984strange}.

\paragraph{Scale and structure.} Our extensible collection currently comprises $131$ previously-published and named chaotic dynamical systems. Each record includes a compilable implementation of the system, a citation reference, default initial conditions on the attractor, precomputed train and test trajectories from different initial conditions at both coarse and fine granularities, and an optimal integration timestep and dominant timescale (used for aligning timescales across systems). For each of the $131$ systems, we include $16$ precomputed trajectories corresponding to all combinations of the following variations per system: coarse and fine sampling granularity, train and test splits emanating from different initial conditions, multivariate and univariate views, and trajectories with and without Brownian noise influencing the dynamics. Because certain data-driven modelling methods, such as our symbolic regression task below, require gradient information, we also include with each system precomputed train and test regression datasets corresponding to trajectories and time derivatives along them.

Figure S1 shows the attractors for all systems, and Table S1 includes brief summaries of their origin and applications. While there are an infinite number of possible chaotic dynamical systems, our work represents, to our knowledge, the first effort to survey and reproduce previously-published chaotic systems. For this reason, while our dataset is readily extensible to new systems, the primary bottleneck as we expand our database is the need to manually reproduce claimed chaotic dynamics, and to identify appropriate parameter values and initial conditions based on published reports. Broadly, our work can be considered a systematization of previous studies that benchmark methods on single chaotic systems such as the Lorenz attractor \cite{nassar2018tree,champion2019data,costa2019adaptive,gilpin2020deep,greydanus2019hamiltonian,lu2020supervised,yu2017long,lu2017reservoir,bellot2021consistency,wang2021reconstructing,li2020scalable,ma2007chaotic}. 

\paragraph{Annotations.} For each system, we calculate and include precise estimates of several standard mathematical characteristics of chaotic systems. More detailed definitions are included in the appendix.\smallskip\\
\textit{The largest Lyapunov exponent} measures the degree to which nearby trajectories diverge, a common measure of the degree of chaos present in a system.\smallskip\\
 \textit{The Lyapunov exponent spectrum} determines the tendency of trajectories to locally converge or diverge across the attractor. All continuous-time chaotic systems have at least one positive exponent, exactly one zero exponent (due to time translation), and, for dissipative systems (i.e., those converging to an attractor), at least one negative exponent \cite{sommerer1993particles}. \smallskip\\ 
 \textit{The correlation dimension} measures an attractor's effective fractal dimension, which informally indicates the intricacy of its geometric structure  \cite{grassberger1983measuring,grebogi1987chaos}. Integer fractal dimensions indicate familiar geometric forms: a line has dimension one, a plane has two, and a filled solid three. Non-integer values correspond to fractals that fill space in a manner intermediate to the two nearest integers. \smallskip\\
 \textit{The multiscale entropy} represents the degree to which complex dynamics persist across timescales \cite{costa2002multiscale}. Chaotic systems have continuous power spectra, and thus high multiscale entropy. \smallskip\\
 We also include two quantities derived from the Lyapunov spectrum: \textit{the Pesin entropy bound}, and \textit{the Kaplan-Yorke fractal dimension}, an alternative estimator of attractor dimension based on trajectory dispersion. Each system is also annotated with various qualitative details, such as whether the system is Hamiltonian or dissipative (i.e., whether there exists conserved invariants like total energy, or whether the dynamics relax to an attractor), non-autonomous (whether the dynamical equations explicitly depend on time), bounded (all variables remain finite as time passes), and whether the dynamics are given by a delay differential equation. In addition to the $131$ differential equations described here, our collection also includes several common discrete time maps; however, we exclude these from our study due to their unique properties.

\paragraph{Methods.} Our dataset includes utilities for re-sampling and re-integrating each system with or without stochasticity, loading pre-computed multivariate or univariate trajectories, computing statistical properties and performing surrogate significance testing, and running benchmarks. One shortcoming of previous studies using chaotic systems as benchmarks---as well as more generally with static time series databases---is inconsistent timescales and granularities (sampling rates). We alleviate this problem by using phase surrogate significance testing to select optimal integration timesteps and sampling rates for all systems in our dataset, thus ensuring that dynamics are aligned across systems with respect to dominant and minimum significant timescales \cite{kantz2004nonlinear}. We further ensure consistency across systems using several standard methods, such as testing ergodicity to find consistent initial conditions, and integrating with continuous re-orthonormalization when computing various mathematical quantities such as Lyapunov exponents (see supplementary material).

\paragraph{Properties and Characterization.} In order to characterize the properties of our collection, we use an off-the-shelf time series featurizer that computes a corpus of $787$ common time series features (e.g. absolute change, peak count, wavelet transform coefficients, etc) for each system in our dataset \cite{christ2018time}. In addition to providing general statistical descriptors for our systems, embedding and clustering the systems based on these features illustrates the diverse dynamics present across our dataset (Figure \ref{fig_stats}). We find that the dynamical systems naturally separate into groups displaying different types of chaotic dynamics, such as smooth scroll-like trajectories versus spiking. Additionally, we observe that our chaotic systems trace a filamentary manifold in embedding space, a property consistent with the established rarity of chaotic attractors within the space of possible dynamical systems: persistent chaos often occurs in an intermediate regime between bifurcations producing simpler dynamics, such as limit cycles or quiescence at fixed points \cite{grebogi1986critical,ott2002chaos}. 
 
%

\subsection{Prior Work.} 

\paragraph{Data-driven modelling and control.}  Many techniques at the intersection of machine learning and dynamical systems theory have been evaluated on specific well-known chaotic attractors, such as the Lorenz, R\"ossler, double pendulum, and Chua systems \cite{nassar2018tree,champion2019data,costa2019adaptive,gilpin2020deep,greydanus2019hamiltonian,lu2020supervised,yu2017long,lu2017reservoir,bellot2021consistency,wang2021reconstructing,li2020scalable,ma2007chaotic}. These and several other chaotic systems used in previous machine learning studies are all included within our dataset \cite{myers2020teaspoon,datseris2018dynamicalsystems}. 
General databases of analytical mathematical models include the BioModels database of systems biology models, which currently contains $1017$ curated entries, with an additional 1271 unreviewed user submissions \cite{le2006biomodels}. Among these models, a subset corresponding to $491$ differential equations appear within the ODEBase database \cite{luders2019odebase}. For the specific task of symbolic regression, the inference of analytical equations from data, existing benchmarks include the Nguyen dataset of $12$ complex mathematical expressions \cite{uy2011semantically}, and corpora of equations from two physics textbooks \cite{udrescu2020ai,orzechowski2018we,strogatz2018nonlinear}, and a recently-released suite of $252$ regression problems from Penn Machine Learning Benchmark \cite{la2021contemporary}. 
\paragraph{Forecasting and classification of time series.} The UCR-UEA time series classification benchmark includes $128$ univariate and $30$ multivariate time series with \textasciitilde$10^1$--$10^3$ timepoints \cite{dau2019ucr,bagnall2017great,dempster2020rocket,bagnall2018uea}. Several of these entries overlap with the UCI Machine Learning Repository, which contains $121$ time series ($91$ multivariate) of lengths \textasciitilde$10^1$--$10^6$ \cite{asuncion2007uci}. The M-series of time series forecasting competitions have most recently featured $10^6$ univariate time series of length \textasciitilde$10^1$--$10^5$ \cite{makridakis2020m4}. The recently-introduced Monash forecasting archive comprises $26$ domain areas, each of which includes \textasciitilde$10^1$--$10^6$ distinct time series with lengths in the range \textasciitilde$10^2$--$10^6$ timepoints  \cite{godahewa2021monash}. A recent long-sequence forecasting model uses the \texttt{ETT-small} dataset of electricity consumption in two regions of China (70,080 datapoints at one-minute increments) \cite{haoyietal2021informer}, as well as NOAA local climatological data (\textasciitilde$10^6$ hourly recordings from \textasciitilde$10^3$ locations) \cite{young2018international}. The PhysioNet database contains several hundred physiological recordings such as EEG, ECG, and blood pressure, at a wide variety of resolutions and lengths \cite{goldberger2000physiobank}.

A point of differentiation between our work and existing datasets is our focus on reproducible chaotic dynamics, which sufficiently narrows the space of potential systems that we can manually curate and re-implement reported dynamics, and calculate key mathematical properties relevant to forecasting and physics-based model inference. These mathematical properties can be used to interpret the properties of black box models by examining their correlation with model performance across systems. Our dataset's curation also ensures a high degree of standardization across systems, such as consistent integration and sampling timescales, as well as ergodicity and stationarity. Additionally, the precomputed multivariate time series in our dataset approximately match the length and size of existing time series databases. We emphasize that, unlike existing time series databases, our dataset's size is flexible due to the ability to re-integrate each system at arbitrary length, sample at any granularity, integrate from new initial conditions, change the amount of stochastic forcing, or even perturb parameters in the underlying differential equation in order to modify or control each system's dynamics.
%

\section{Experiments}


\subsection*{Task 1: Forecasting}

\begin{figure}
  \centering
 \includegraphics[width=\linewidth]{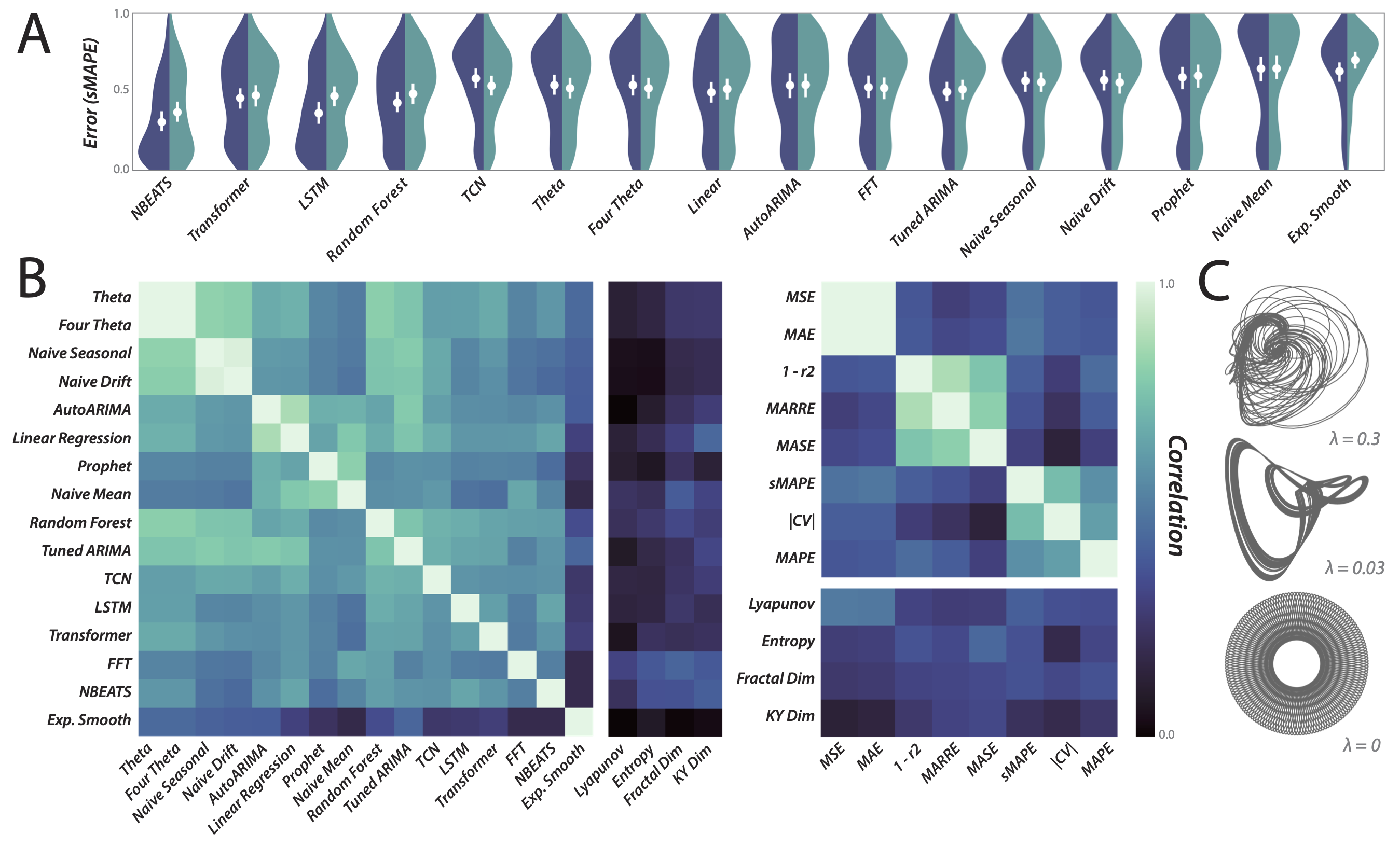}
  \caption{{\bf Forecasting benchmarks for all chaotic dynamical systems.} (A) Distribution of forecast errors for all dynamical systems and for all forecasting models, sorted by increasing median error. Dark and light hues correspond to coarse and fine time series granularities.  (B) Spearman correlation among forecasting models, among different forecast evaluation metrics, and between forecasting metrics and underlying mathematical properties, computed across all dynamical systems at fine granularity. Columns are ordered by descending maximum cross-correlation in order to group similar models and metrics. (C) The systems with the highest, median, and lowest forecasting error across all models, annotated by largest Lyapunov exponent.}
  \label{forecast}
  \vspace{-3mm}
\end{figure}

Chaotic systems are inherently unpredictable, and extensive work by the physics community has sought to quantify chaos, and to relate its properties to general features of the underlying governing equations \cite{ott2002chaos,tang2020introduction}. Traditionally, the predictability of a chaotic system is thought to be determined by the largest Lyapunov exponent, which measures the rate at which trajectories emanating from two infinitesimally-spaced points will exponentially separate over time \cite{sommerer1993particles}. 

We evaluate this claim on our dataset by benchmarking $16$ forecasting models spanning a wide variety of techniques: deep learning methods (NBEATS, Transformer, LSTM, and Temporal Convolutional Network), statistical methods (Prophet, Exponential Smoothing, Theta, 4Theta), common machine learning techniques (Random Forest), classical methods (ARIMA, AutoARIMA, Fourier transform regression), and standard naive baselines (naive mean, naive seasonal, naive drift) \cite{oreshkin2019n,lea2016temporal,alexandrov2020gluonts,godahewa2021monash}. Our train and test datasets correspond to differential initial conditions, and we perform separate hyperparameter tuning for each chaotic system and granularity \cite{unit8co2020darts,alexandrov2020gluonts}. While the forecasting models are heterogenous, for each we tune whichever hyperparameter most closely corresponds to a timescale---for example, the lag order for autoregressive models, or the input chunk size for the neural network models. Because all systems are aligned to the same average period, the range of values over which timescales are tuned is scaled by the granularity. Hyperparameters are tuned using held-out future values, and scores are computed on an unseen test trajectory emanating from different initial conditions.

Our results are shown in Figure \ref{forecast} for all dynamical systems at coarse and fine sampling granularity. We include corresponding results for systems with noise in the supplementary material. We find the the deep learning models perform particularly well, with the Transformer and NBEATS models achieving the lowest median scores, while also appearing within the three best-performing models for nearly all systems. On many datasets, the temporal convolutional network and traditional LSTM models also achieve competitive performance. Notably, the random forest also exhibits strong performance despite the continuous nature of our datasets, and with substantially lower training cost. 
The relative ranking of the different forecasting models remains stable both as granularity is varied over two orders of magnitude, and as noise is increased to a level dominating the signal (see supplementary experiments). In the latter case, we observe that the performance of different models converges as their overall performance decreases. Overall, NBEATS strongly outperforms the other forecasting techniques across varied systems and granularities, and its performance persists even in the presence of noise. We speculate that NBEAT's advantage arises from its implicit decomposition of time series into a hierarchy of basis functions \cite{oreshkin2019n}, an approach that mirrors classical techniques for representing continuous-time chaotic systems \cite{wang2011predicting}.

Our results seemingly contrast with studies showing that statistical models outperform neural networks on forecasting tasks \cite{godahewa2021monash,makridakis2020m4}. However, our forecasting task focuses on long time series and prediction horizons, two areas where neural networks have previously performed well \cite{haoyietal2021informer}. Additionally, we hypothesize that the strong performance of deep learning models on our dataset is a consequence of the smoothness of chaotic systems, which have mathematical regularity and stationarity compared to time series generated from industrial or environmental measurements. In contrast, models like Prophet are often applied to calendar data with seasonality and irregularities like holidays \cite{taylor2018forecasting}---neither of which have a direct analogue in chaotic systems, which contain a continuous spectrum of frequencies \cite{lusch2018deep}. Consistent with this intuition, we observe that among the systems in our dataset, the Prophet model performs well on the torus, a quasiperiodic system with largest Lyapunov exponent equal to zero.

Several recent works have considered the appropriate metric for determining forecast accuracy \cite{hyndman2006another,makridakis2020m4,durbin2012time,godahewa2021monash}. For all forecasting models and dynamical systems we compute eight error metrics: the mean squared error (MSE), mean absolute scaled error (MASE), mean absolute error (MAE), mean absolute ranged relative error (MARRE), the magnitude of the coefficient of variation ($\abs{CV}$), one minus the coefficient of determination ($1 - r^2$), and the symmetric and regular mean absolute percent errors (MAPE and sMAPE). We find that all of these potential metrics are positively correlated across our dataset, and that they can be grouped into families of strongly-related metrics (Figure \ref{forecast}B). We also observe that the relative ranking of different forecasting models is independent of the choice of metric. Hereafter, we report sMAPE errors when comparing models, but we include all other metrics within the benchmark.

We next evaluate the common claim that the empirical predictability of a system depends on the mathematical degree of chaos present \cite{pathak2018model}. For each system, we correlate the forecast error of the best-performing model with the various mathematical properties of each system (Figure \ref{forecast}B). Across all systems, we find a high degree of correlation between the largest Lyapunov exponent and the forecast error, while other measures such as the attractor fractal dimension and entropy correlate less strongly. While this observation matches conventional wisdom, it represents (to our knowledge) the first large-scale test of the empirical relevance of Lyapunov exponents. We consider this observation particularly noteworthy because our forecasting task spans several periods, yet the Lyapunov exponent measures only the local dispersion between infinitesimally-separated points.

Our results introduce several considerations for the development of time series models. The strong performance we observe for neural network models implies that the flexibility of large models proves beneficial for time series without obvious trends or seasonality. The consistent accuracy we observe for NBEATS, even in the presence of noise, suggests that hierarchical decomposition can improve modelling of systems with multiple timescales. Most of our best-performing methods implicitly lift the dimensionality of the input time series, implying that higher-dimensional representations create more predictable dynamics—a finding consistent with recent studies showing that certain machine learning techniques implicitly learn Koopman operators, linear propagators that act on lifted representations of nonlinear systems \cite{lusch2018deep,champion2019data,klus2018data,otto2021koopman,takeishi2017learning}. That higher dimensional representations can linearize dynamics mirrors classical motivation for kernel methods in machine learning \cite{hastie2009elements}; we thus hypothesize that classical time series representations like time-lagged embeddings can be improved through nonlinearities, either in the form of custom functions learned by neural networks, or by inductive biases in the form of fixed periodic or wavelet-like kernels.

\subsection*{Task 2: Accelerating model training with importance sampling.}
%
%
%
%
 \begin{table}
  \caption{(Upper) Forecast accuracy for LSTMs trained on full time series, random subsets, and subsets sampled proportionately to their epochwise error (medians $\pm$ standard errors across all dynamical systems). (Lower) Accuracy scores on the UCR database for classifiers trained on features extracted from bare time series, and from autoencoders pretrained on the full chaotic systems collection at random and task-matched timescales (medians $\pm$ standard errors across UCR tasks).}
  \label{importance}
  \centering
  \begin{tabular}{llll}
\bottomrule
    \toprule
    \multicolumn{2}{c}{Importance Sampling Forecasting Error (sMAPE)}                   \\
    \cmidrule(r){1-2}
\null &    Full Epochs     & Random Subset     & Importance Weighted \\
\midrule
 sMAPE & $1.00 \pm 0.05$& $0.99 \pm 0.05$ & $\mathbf{0.90 \pm 0.05}$     \\
 Runtime (sec) & $190.1 \pm 0.3$&  $\mathbf{77.9 \pm 0.3}$ &  $94.6 \pm 0.2$    \\
    \bottomrule
    \toprule
    \multicolumn{2}{c}{Transfer Learning Classification Accuracy}                   \\
    \cmidrule(r){1-2}
\null &    No Transfer Learning    & Random Timescales     & Matched Surrogates \\
    \midrule
  Accuracy &  $0.80 \pm 0.02$& $0.82 \pm 0.01$ & $\mathbf{0.84 \pm 0.01}$     \\
   \bottomrule\toprule
  \end{tabular}
  \vspace{-5mm}
\end{table}

When training a forecasting model iteratively, each training batch usually samples input timepoints with equal probability. However, chaotic attractors generally possess non-uniform measure due to their fractal structure \cite{farmer1982dimension}. We thus hypothesize that importance sampling can accelerate training of a forecast model, by encouraging the network to oversample sparser regions of the underlying attractor \cite{leitao2013monte}. We thus modify the training procedure for a forecast model by applying a simple form of importance sampling, based on the epoch-wise training losses of individual samples---an approach related to zeroth-order adaptive methods appearing in other areas \cite{press1986numerical,jiang2019accelerating,kawaguchi2020ordered,katharopoulos2018not}. Our procedure consists of the following: (1) We halt training every few epochs and compute historical forecasts (backtests) on the training trajectory. (2) We randomly sample timepoints proportionately to their error in the historical forecast, and then generate a set of initial conditions corresponding to random perturbations away from each sampled attractor point. (3) We simulate the full dynamical system for $\tau$ timesteps for each of these initial conditions, and we use these new trajectories as the training set for the next $b$ epochs. We repeat this procedure for $\nu$ meta-epochs. For the original training procedure, the training time scales as $\sim B$, the number of training epochs. In our modified procedure, the training time has dominant term $\sim \nu \,b$, plus an additional term proportional to $\tau$ (integration can be parallelized across initial conditions), plus a small constant cost for sampling. We thus set $\nu \, b < B$, and record run times to verify that total cost has decreased.

Table \ref{importance} shows the results of our experiments for an LSTM model across all chaotic attractors. Importance sampling achieves a significantly smaller forecast error than a baseline using the full training set in each epoch, as well as a control in which the exact importance sampling procedure was repeated without weighting random samples by error (two sided paired t-test, $p < 10^{-6}$ for both tests). Notably, importance sampling requires substantially lower computation due to the reduced number of training epochs incurred. Our approach exploits that our database comprises strange {\it attractors}, because initial conditions derived from random perturbations off an attractor will produce trajectories that return to the attractor.

\subsection*{Task 3: Transfer learning and data augmentation.}

We next explore how our dataset can assist general time series analysis, regardless of the relevance of chaos to the problem. We study an existing time series classification benchmark, and we use our dataset to generate timescale-matched surrogate data for transfer learning. 

Our classification procedure broadly consists of training an autoencoder on trajectories from our database, and then using the trained encoder as a general feature extractor for time series classification. However, unlike existing transfer learning approaches for time series \cite{malhotra2017timenet}, we train the autoencoder on a new dataset for each classification problem: we re-integrate our entire dataset to match the dominant timescales in the classification problem's training data.

Our approach thus comprises several steps: (1) Across all data in the train partition, the dominant significant Fourier frequency is determined using random phase surrogates \cite{kantz2004nonlinear}. (2) Trajectories are re-integrated for every dynamical system in our database, such that the sampling rate of the dynamics is equal to that of the training dataset. The surrogate ensemble thus corresponds to a custom set of trajectories with timescales matched to the training data of the classification problem. (3) We train an autoencoder on this ensemble. Our encoder is a one layer causal dilated encoder with skip connections, an architecture recently shown to provide strong time series classification performance \cite{franceschi2019unsupervised}. (3) We apply the encoder to the training data of the classification problem. (4) We apply a standard linear time series classifier, following recent works \cite{dempster2020rocket,loning2019sktime}. We featurize the time series using a library of standard featurizers \cite{christ2018time}, and then perform classification using ridge regression \cite{loning2019sktime}. Overall, our classification approach bears conceptual similarity to other generative data augmentation techniques: we extract parameters (the dominant timescales) from the training data, and then use these parameters to construct a custom surrogate ensemble with matching timescales. In many image augmentation approaches, a prior distribution is learned from the training data (e.g. via a GAN), and the then sampled to create surrogate examples \cite{zhang2019dada,tran2017bayesian,zhu2017data,hauberg2016dreaming}.

\begin{wrapfigure}{r}{0.4\linewidth}
\vspace{-5mm}
  \centering 
 \includegraphics[width=\linewidth]{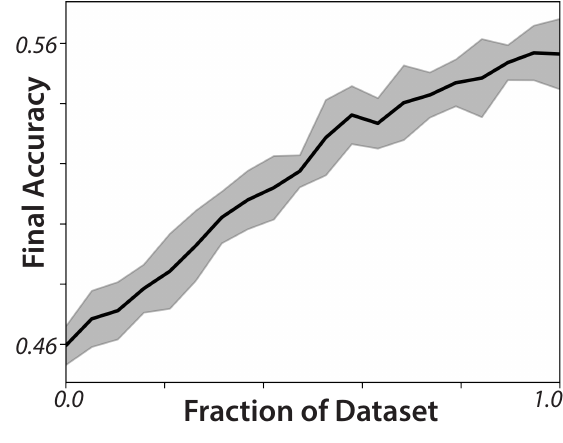}
  \caption{Classification accuracy on the UCR dataset \texttt{EOGHorizontalSignal}, across models pretrained on increasing fractions of the database. Standard errors are from bootstrapped replicates, where the dynamical systems are sampled with replacement.}
\label{transfer}
\vspace{-5mm}
\end{wrapfigure}

As baselines for our approach, we train a classifier on the bare original time series, as well as a "random timescale" collection in which the time series in the surrogate ensemble have random dominant frequencies, unrelated to the timescales in the training data. The latter ablation serves to isolate the role of timescale matching, which is uniquely enabled by the ability to re-integrate our dataset at arbitrary granularity. This control experiment is necessary in light of recent work showing that transfer learning on a large collection of time series can yield informative features \cite{malhotra2017timenet}.

We benchmark classification using the UCR time series classification benchmark, which contains $128$ real-world classification problems spanning diverse areas like medicine, agriculture, and robotics  \cite{dau2019ucr}. Because we are using convolutional models, we restrict our analysis to the $91$ datasets with at least $100$ contiguous timepoints (these include the $85$ "bakeoff" datasets benchmarked in previous studies) \cite{bagnall2017great}. We compute separate benchmarks (surrogate ensembles, features, and scores) for each dataset in the archive.

Our results are shown in Table \ref{importance}. Across the UCR archive we observe statistically significant average classification accuracy increases of $4\% \pm 1\%$ compared to the raw dataset ($p < 10^{-4}$, paired two-sided t-test), and $2\% \pm 1\%$ compared to the ablation with random surrogate timescales ($p < 10^{-4}$). While these modest improvements do not comprise state-of-the-art results on the UCR database \cite{bagnall2017great}, they demonstrate that features learned from chaotic systems in an unsupervised setting can be used to extract meaningful general features for further analysis. On certain datasets, our results approach other recent unsupervised approaches in which a simple linear classifier is trained on top of a complex unsupervised feature extractor \cite{malhotra2017timenet,franceschi2019unsupervised,dempster2020rocket}. Recent results have even shown that a very large number of {\it random} convolutional features can provide informative representations of time series for downstream supervised learning tasks \cite{dempster2020rocket}; we therefore speculate that pretraining with chaotic systems may allow more efficient selection of informative convolutional kernels. Moreover, the improvement of transfer learning over the random timescale model demonstrates the advantage of re-integration. In order to verify that the diversity of dynamical systems present within our dataset contribute to the quality of the learned features, we repeat the classification task on a single UCR dataset, corresponding to clinical eye tracking data. We train encoders on gradually increasing numbers of dynamical systems, in order to see how the final accuracy changes as the number of systems available for pretraining increases (Figure \ref{transfer}). We observe monotonic scaling, indicating that our dataset's size and diversity contribute to feature quality.

\subsection*{Task 4: Data-driven model inference and symbolic regression}

We next look beyond traditional time series analysis, and apply our database to a data-driven modelling task. A growing body of work uses machine learning methods to infer dynamical systems directly from data \cite{karniadakis2021physics,de2020discovery,callaham2021learning,carleo2019machine}. Examples include constructing effective propagators for the dynamics \cite{otto2021koopman,costa2021maximally,takeishi2017learning,budivsic2012applied,gilpin2019cellular,lusch2018deep}, obtaining neural representations of the dynamical equations \cite{chen2018neural,kidger2020neural,massaroli2020dissecting,rackauckas2020universal}, and inferring analytical governing equations via symbolic regression \cite{klus2018data,petersen2019deep,brunton2016discovering,schmidt2009distilling,martin2018reverse,udrescu2020ai,cranmer2020discovering,la2021contemporary,rudy2021sparse}. Beyond improving forecasts, these data-driven representations can discover mechanistic insights, such as symmetries or separated timescales, that might not otherwise be apparent in a time series. 

\begin{figure}
  \centering
 \includegraphics[width=0.6\linewidth]{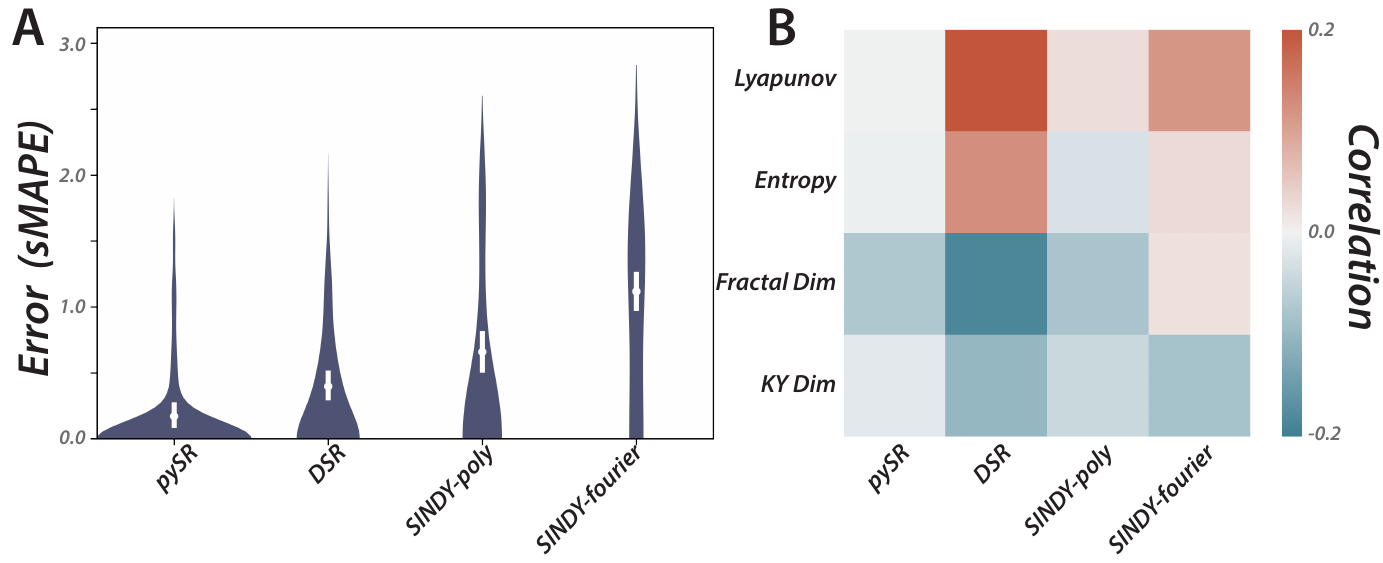}
  \caption{{\bf Symbolic regression benchmarks.} (A) Error distributions on test datasets across all systems, (B) Spearman correlation between errors and mathematical properties of the underlying systems.}
\label{fig_symb}
 \vspace{-4mm}
\end{figure}

We thus use our dataset for data-driven modelling in the form of a symbolic regression task. We focus on symbolic regression because of the recent emergence of widely-used benchmark models and performance desiderata for these methods \cite{la2021contemporary}. However, we emphasize that our database can be used for other emerging focus areas in data-driven modelling, such as inference of empirical propagators or neural ordinary differential equations \cite{lusch2018deep,froyland2009almost,budivsic2012applied}, and we include a baseline neural ordinary differential equation task in the supplementary material. For each dynamical system in our collection, we generate trajectories with sufficiently coarse granularity to sample all regions of the attractor. At each timepoint, we compute the value of the right hand side of the corresponding dynamical equation, and we treat the value of this time derivative as the regression target. We use this dataset to compare several recent symbolic regression approaches: (1) DSR: a recurrent neural network trained with a risk-seeking policy gradient, which produces state-of-the-art results on a variety of challenging symbolic regression tasks \cite{petersen2019deep}. (2) PySR: an open-source package inspired by the popular closed-source software Eureqa, which uses genetic programming and simulated annealing \cite{cranmer2020pysr,cranmer2020discovering,schmidt2009distilling}. (3,4) PySINDY: a Python implementation of the widely-used SINDY algorithm, which uses sparse regression to decompose data into linear combinations of functions \cite{de2020pysindy,brunton2016discovering}. For PySINDY we train separate models for purely polynomial (SINDY-poly) and trigonometric (SINDY-fourier) bases. For DSR and pySR we use a standard library of binary and unary expressions, $\{+, - , \times, \div \}$, $\{\sin, \cos, \exp, \log, \tanh \}$ \cite{petersen2019deep}. After fitting a formula using each method, we evaluate it on an unseen test trajectory arising from different initial conditions, and we report the the sMAPE error between the formula's prediction and the true value along the trajectory.

Our results illustrate several features of our dataset, while also illustrating properties of the different symbolic regression algorithms. All algorithms show strong performance across the chaotic systems dataset (Figure \ref{fig_symb}). The two lowest-error models, pySR and DSR, exhibit nearly-equivalent performance when accounting for error ranges, and both achieve errors near zero on many systems. We attribute this strong performance to the relatively simple algebraic construction of most published systems: named chaotic systems will inevitably favor concise, demonstrative equations over complex expressions. In fact, several systems in our dataset belong to the Sprott attractor family, which represent the algebraically simplest chaotic systems \cite{sprott1994some}. In this sense, our dataset likely has similar complexity to the Feynman equations benchmark \cite{udrescu2020ai}.

We highlight that PySINDY with a purely polynomial basis performs very well on our dataset, especially as a linear approach that requires a median training time of only $0.01 \pm 0.01$ s per system on a single CPU core. In comparison, pySR had a median time of $1400 \pm 60$ s per system on one core, while DSR on one GPU required $4300 \pm 200$s per system---consistent with the results of recent symbolic regression benchmark suite \cite{la2021contemporary}. However, parallelization reduces the runtime of all methods.

We emphasize that the relative performance of a given symbolic regression algorithm depends on diverse factors, such as equation complexity, the library of available unary and binary operators, the amount and dynamic range of available input data, the amount of compute available for refinement, and the degree of nonlinearity of the underlying system. More generally, symbolic regression algorithms exhibit a bias-variance tradeoff manifesting as Pareto front bridging accuracy and parsimony: large models with many terms will appear more accurate, but at the expense of brevity and potentially robustness and interpretability \cite{schmidt2009distilling}. More challenging benchmarks would include nested expressions and uncommon transcendental functions; these systems may be a more appropriate setting for benchmarking state-of-the-art techniques like DSR. Additionally, we do not include measurement noise in our experiments, a scenario in which DSR performs strongly compared to other methods \cite{petersen2019deep,la2021contemporary}.

Interestingly, DSR exhibits the strongest dependence on the mathematical properties of the underlying dynamics: more chaotic systems consistently yield higher errors (Figure \ref{fig_symb}B). We consider this result surprising, because {\it a priori} we would expect the performance of a given symbolic regression algorithm to depend purely on the syntactic complexity of the target formula, rather than the dynamics that it produces. Because DSR uses a large model to navigate a space of smaller models, we hypothesize that more chaotic systems present a broader set of possible "partial formulae" that match specific subregimes of the attractor---an effect exploited in several recent decomposition techniques for chaotic systems \cite{nassar2018tree,costa2019adaptive}. The diversity of these local approximants would result in a more complex global search space.

\section{Discussion}

We have introduced an extensible collection of known chaotic dynamical systems. In addition to representing a customizable benchmark for time series analysis and data-driven modelling, we have provided examples of additional applications, such as transfer learning for general time series analysis tasks, that are enabled by the generative nature of our dataset. We note that there are several other potential applications that we have not explored here: testing feedback-based control algorithms (which require perturbing the parameters of a given dynamical system, and then re-integrating), and inferring numerical propagators (such as Koopman operators)\cite{otto2021koopman,costa2021maximally,takeishi2017learning,budivsic2012applied,gilpin2019cellular,lusch2018deep,gilpin2020learning,arbabi2017ergodic}. In the appendix, we include preliminary benchmarks for a neural ordinary differential equations task \cite{chen2018neural,kidger2020neural,massaroli2020dissecting,rackauckas2020universal}; due to the direct connections between our work and this area, we hope to further explore these methods in future studies. Our work can be seen as systematizing the common practice of testing new methods on single chaotic systems, particularly the Lorenz attractor \cite{nassar2018tree,champion2019data,costa2019adaptive,gilpin2020deep,greydanus2019hamiltonian,lu2020supervised,yu2017long,lu2017reservoir,bellot2021consistency,wang2021reconstructing,li2020scalable,ma2007chaotic}.

More broadly, our collection seeks to improve the interpretability of data-driven modelling from time series. For example, our forecasting benchmark experiments show that the Lyapunov exponent, a popular measure of chaoticity, correlates with the empirical predictability of a system under a variety of models---a finding that matches intuition, but which has not (to our knowledge) previously been tested extensively. Likewise, in our symbolic regression benchmark we find that more chaotic systems are harder to model, an effect we attribute to the diverse local approximants available for complex dynamical systems. These examples demonstrate how the control and mathematical context provided by differential equations can yield mechanistic insight beyond traditional time series.

Limitations of our approach include our inclusion only of known chaotic systems that have previously appeared in published works. This limits the rate at which our collection may expand, since each new entry requires manual curation and implementation in order to verify reported dynamics. Our focus on published systems may bias the dataset towards more unusual (and thus reportable) dynamics, particularly because there are infinite possible chaotic systems. Additionally, our model scoring metrics primarily quantify point-wise forecast accuracy; however, over long timescales it may be informative to consider alternative metrics such as stationarity with respect to dynamical properties, or the accuracy of the topology of the predicted attractor \cite{schmidt2019identifying,gilpin2020deep,koppe2019identifying}. More broadly, we note that in few dimensions chaotic dynamics are rare relative to the space of all possible models \cite{ott2002chaos}, although chaos becomes ubiquitous as the number of coupled variables increases \cite{ispolatov2015chaos}. Nonetheless, low-dimensional chaos may represent an instructive step towards understanding complex dynamics in high-dimensional systems. 


\begin{ack}
We thank Gautam Reddy, Samantha Petti, Brian Matejek, and Yasa Baig for helpful discussions and comments on the manuscript. W. G. was supported by the NSF-Simons Center for Mathematical and Statistical Analysis of Biology at Harvard University, NSF Grant DMS 1764269, and the Harvard FAS Quantitative Biology Initiative. The author declares no competing interests.
\end{ack}

\bibliography{dysts_cites}
\bibliographystyle{naturemag}

\end{document}


\title{Supplementary material for "Chaos as an interpretable benchmark for forecasting and data-driven modelling"}



\renewcommand{\thetable}{S\arabic{table}}
\setcounter{table}{0}
\renewcommand{\thefigure}{S\arabic{figure}} 
\setcounter{figure}{0}
\renewcommand{\theequation}{A\arabic{equation}}
\setcounter{equation}{0}
\renewcommand{\thesubsection}{\Alph{subsection}}
\setcounter{subsection}{0}

\maketitle
\tableofcontents

\section{Data Availability}

The database of dynamical models and precomputed time series is available on GitHub at \url{https://github.com/williamgilpin/dysts}. The \texttt{benchmarks} subdirectory contains all code needed reproduce the benchmarks, figures, and tables in this paper.

All included equations are in the public domain, and all precomputed time series datasets have been generated {\it de novo} from these equations. No license is required to use these equations or datasets. The repository and precomputed datasets include an Apache 2.0 license. The author attests that they bear responsibility for copyright matters associated with this dataset. 

\clearpage

\begin{figure*}
  \centering
 \includegraphics[width=\linewidth]{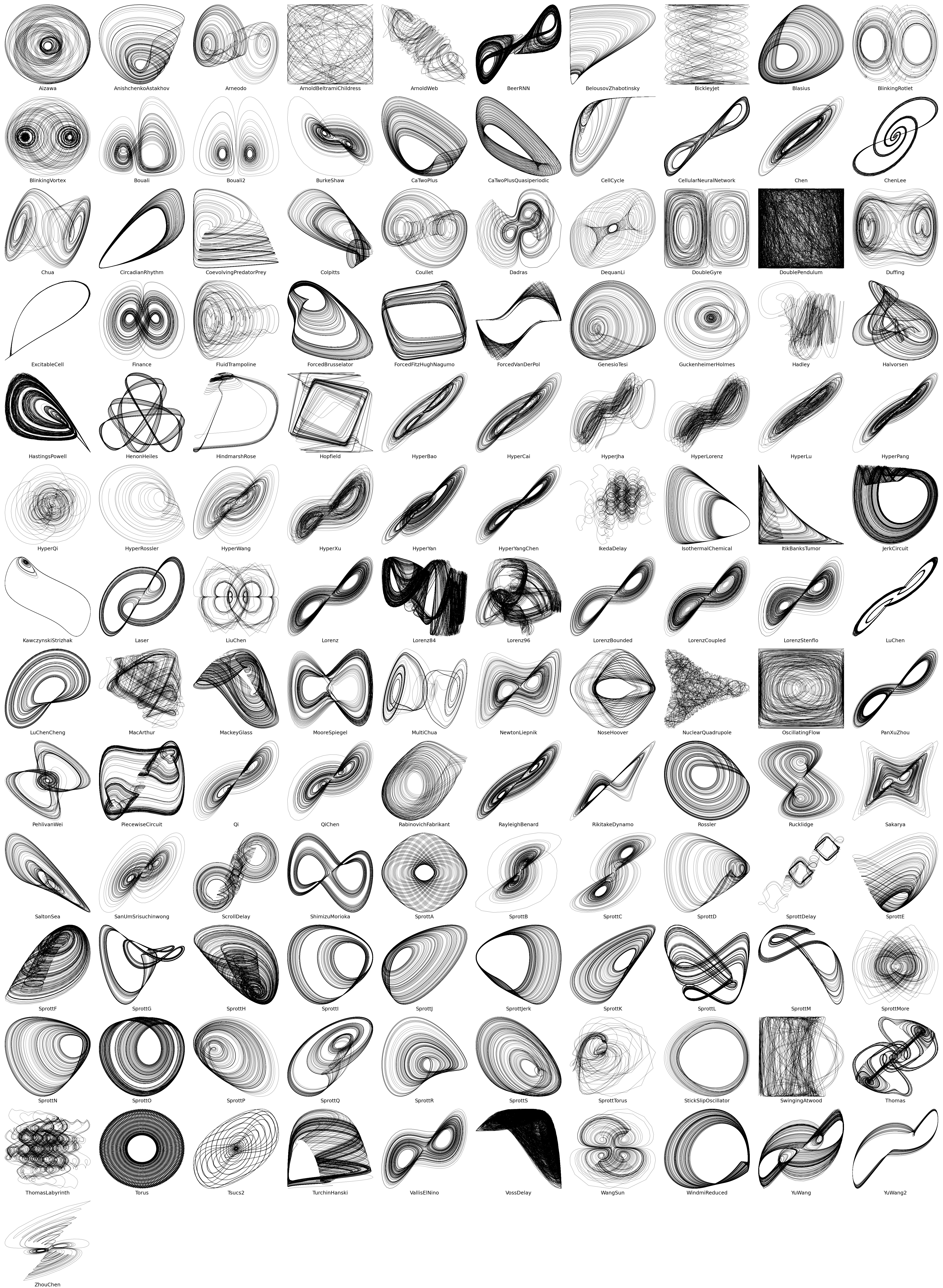}
  \caption{{\bf All dynamical systems currently in the database.}
 }
 \label{fig_all}
\end{figure*}

\clearpage

\section{Descriptions of all systems}
 
Descriptions and citations for all systems are included below, and each system is visualized in Figure \ref{fig_all}. Each system's entry in the project repository contains full records and descriptions.

\setlength\LTleft{-2cm}
\begin{longtable}{lll}
\toprule
System &                                           Reference &                                        Description \\
\toprule
Aizawa                  &  Aizawa, Yoji, and Tatsuya Uezu (1982). Topolog... &  A torus-like attractor related to the forced L... \\
AnishchenkoAstakhov     &  Anishchenko, et al. Nonlinear dynamics of chao... &        Stochastic resonance in forced oscillators. \\
Arneodo                 &  Arneodo, A., Coullet, P. \& Tresser, C. Occuren... &  A modified Lotka-Volterra ecosystem, also know... \\
ArnoldBeltramiChildress &  V. I. Arnold, Journal of Applied Mathematics a... &  An exact solution of Euler's equation for invi... \\
ArnoldWeb               &  Froeschle, C., Guzzo, M. \& Legga, E (2000). Gr... &  A quasi-integrable system that transitions to ... \\
BeerRNN                 &  Beer, R. D. (1995). On the dynamics of small c... &         A two-neuron minimal model nervous system. \\
BelousovZhabotinsky     &  Gyorgyi and Field (1992). A three-variable mod... &  A reduced-order model of the BZ reaction that ... \\
BickleyJet              &  Hadjighasem, Karrasch, Teramoto, Haller (2016)... &  A zonal jet passing between two counter rotati... \\
Blasius                 &               Blasius, Huppert, Stone. Nature 1999 &   A chaotic food web composed of interacting pr... \\
BlinkingRotlet          &  Meleshko \& Aref. A blinking rotlet model for c... &  The location of the mixer is chosen so that th... \\
BlinkingVortex          &  Aref (1984). Stirring by chaotic advection. J.... &  A classic minimal chaotic mixing flow. Solutio... \\
Bouali                  &  Bouali (1999). Feedback loop in extended Van d... &  Economic cycles with fluctuating demand. Relat... \\
Bouali2                 &  Bouali (1999). Feedback loop in extended Van d... &                   A modified economic cycle model. \\
BurkeShaw               &       Shaw (1981). Zeitschrift fur Naturforschung. &  A scroll-like attractor with unique symmetry a... \\
CaTwoPlus               &    Houart, Dupont, Goldbeter. Bull Math Biol 1999. &            Intracellular calcium ion oscillations. \\
CaTwoPlusQuasiperiodic  &    Houart, Dupont, Goldbeter. Bull Math Biol 1999. &  Intracellular calcium ion oscillations with qu... \\
CellCycle               &           Romond, Rustici, Gonze, Goldbeter. 1999. &  A simplified model of the cell cycle. The para... \\
CellularNeuralNetwork   &  Arena, Caponetto, Fortuna, and Porto., Int J B... &                  Cellular neural network dynamics. \\
Chen                    &  Chen (1997). Proc. First Int. Conf. Control of... &  A system based on feedback anti-control in eng... \\
ChenLee                 &  Chen HK, Lee CI (2004). Anti-control of chaos ... &           A rigid body with feedback anti-control. \\
Chua                    &  Chua, L. O. (1969) Introduction to Nonlinear N... &  An electronic circuit with a diode providing n... \\
CircadianRhythm         &   Leloup, Gonze, Goldbeter. 1999. Gonze, Leloup... &  The Drosophila circadian rhythm under periodic... \\
CoevolvingPredatorPrey  &            Gilpin \& Feldman (2017). PLOS Comp Biol &  A system of predator-prey equations with co-ev... \\
Colpitts                &  Kennedy (2007). IEEE Trans Circuits \& Systems.... &  An electrical circuit used as a signal generator. \\
Coullet                 &  Arneodo, A., Coullet, P. \& Tresser, C. Occuren... &                 A variant of the Arneodo attractor \\
Dadras                  &  S Dadras, HR Momeni (2009). A novel three-dime... &  An electronic circuit capable of producing mul... \\
DequanLi                &                    Li, Phys Lett A. 2008: 387-393. &  Related to the Three Scroll unified attractor ... \\
DoubleGyre              &  Shadden, Lekien, Marsden (2005). Definition an... &  A time-dependent fluid flow exhibiting Lagrang... \\
DoublePendulum          &  See, for example: Marion (2013). Classical dyn... &         Two coupled rigid pendula without damping. \\
Duffing                 &  Duffing, G. (1918), Forced oscillations with v... &  A monochromatically-forced rigid pendulum, wit... \\
ExcitableCell           &  Teresa Chay. Chaos In A Three-variable Model O... &  A reduced-order variant of the Hodgkin-Huxley ... \\
Finance                 &  Guoliang Cai, Juanjuan Huang. International Jo... &  Stock fluctuations under varying investment de... \\
FluidTrampoline         &  Gilet, Bush. The fluid trampoline: droplets bo... &      A droplet bouncing on a horizontal soap film. \\
ForcedBrusselator       &  I. Prigogine, From Being to Becoming: Time and... &                  An autocatalytic chemical system. \\
ForcedFitzHughNagumo    &  FitzHugh, Richard (1961). Impulses and Physiol... &  A driven neuron model sustaining both quiesent... \\
ForcedVanDerPol         &  B. van der Pol (1920). A theory of the amplitu... &         An electronic circuit containing a triode. \\
GenesioTesi             &  Genesio, Tesi (1992). Harmonic balance methods... &          A nonlinear control system with feedback. \\
GuckenheimerHolmes      &  Guckenheimer, John, and Philip Holmes (1983). ... &                            A nonlinear oscillator. \\
Hadley                  &  G. Hadley (1735). On the cause of the general ... &                    An atmospheric convective cell. \\
Halvorsen               &  Sprott, Julien C (2010). Elegant chaos: algebr... &  An algebraically-simple chaotic system with qu... \\
HastingsPowell          &                     Hastings, Powell. Ecology 1991 &                          A three species food web. \\
HenonHeiles             &  Henon, M.; Heiles, C. (1964). The applicabilit... &        A star's motion around the galactic center. \\
HindmarshRose           &       Marhl, Perc. Chaos, Solitons, Fractals 2005. &    A neuron model exhibiting spiking and bursting. \\
Hopfield                &                    Lewis \& Glass, Neur Comp (1992) &      A neural network with frustrated connectivity \\
HyperBao                &  Bao, Liu  (2008). A hyperchaotic attractor coi... &                       Hyperchaos in the Lu system. \\
HyperCai                &  Guoliang, Huang (2007). A New Finance Chaotic ... &      A hyperchaotic variant of the Finance system. \\
HyperJha                &  Jürgen Meier (2003). Presentation of Attractor... &                             A hyperchaotic system. \\
HyperLorenz             &  Jürgen Meier (2003). Presentation of Attractor... &    A hyperchaotic variant of the Lorenz attractor. \\
HyperLu                 &  Jürgen Meier (2003). Presentation of Attractor... &        A hyperchaotic variant of the Lu attractor. \\
HyperPang               &  Jürgen Meier (2003). Presentation of Attractor... &                             A hyperchaotic system. \\
HyperQi                 &  G. Qi, M. A. van Wyk, B. J. van Wyk, and G. Ch... &           A hyperchaotic variant of the Qi system. \\
HyperRossler            &  Rossler, O. E. (1979). An equation for hyperch... &      A hyperchaotic variant of the Rossler system. \\
HyperWang               &  Wang, Z., Sun, Y., van Wyk, B. J., Qi, G. \& va... &         A hyperchaotic variant of the Wang system. \\
HyperXu                 &  Letellier \& Rossler (2007). Hyperchaos. Schola... &                             A hyperchaotic system. \\
HyperYan                &  Jürgen Meier (2003). Presentation of Attractor... &                             A hyperchaotic system. \\
HyperYangChen           &  Jürgen Meier (2003). Presentation of Attractor... &                             A hyperchaotic system. \\
IkedaDelay              &  K. Ikeda and K. Matsumoto (1987). High-dimensi... &  A passive optical resonator system. A standard... \\
IsothermalChemical      &  Petrov, Scott, Showalter. Mixed-mode oscillati... &  An isothermal chemical system with mixed-mode ... \\
ItikBanksTumor          &             Itik, Banks. Int J Bifurcat Chaos 2010 &                A model of cancer cell populations. \\
JerkCircuit             &  Sprott (2011). A new chaotic jerk circuit. IEE... &  An electronic circuit with nonlinearity provid... \\
KawczynskiStrizhak      &  P. E. Strizhak and A. L. Kawczynski, J. Phys. ... &  A chemical oscillator model describing mixed-m... \\
Laser                   &  Abooee, Yaghini-Bonabi, Jahed-Motlagh (2013). ... &                        A semiconductor laser model \\
LiuChen                 &  Liu, Chen. Int J Bifurcat Chaos. 2004: 1395-1403. &                              Derived from Sakarya. \\
Lorenz                  &  Lorenz, Edward N (1963). Deterministic nonperi... &  A minimal weather model based on atmospheric c... \\
Lorenz84                &  E. Lorenz (1984). Irregularity: a fundamental ... &  Atmospheric circulation analogous to Hadley co... \\
Lorenz96                &  Lorenz, Edward (1996). Predictability: A probl... &  A climate model containing fluid-like advectiv... \\
LorenzBounded           &                      Sprott \& Xiong (2015). Chaos. &  The Lorenz attractor in the presence of a conf... \\
LorenzCoupled           &  Lorenz, Edward N. Deterministic nonperiodic fl... &                     Two coupled Lorenz attractors. \\
LorenzStenflo           &  Letellier \& Rossler (2007). Hyperchaos. Schola... &                Atmospheric acoustic-gravity waves. \\
LuChen                  &     Lu, Chen. Int J Bifurcat Chaos. 2002: 659-661. &  A system that switches shapes between the Lore... \\
LuChenCheng             &  Lu, Chen, Cheng. Int J Bifurcat Chaos. 2004: 1... &  A four scroll attractor that reduces to Lorenz... \\
MacArthur               &  MacArthur, R. 1969. Species packing, and what ... &  Population abundances in a plankton community,... \\
MackeyGlass             &  Glass, L. and Mackey, M. C. (1979). Pathologic... &  A physiological circuit with time-delayed feed... \\
MooreSpiegel            &  Moore, Spiegel. A Thermally Excited Nonlinear ... &                    A thermo-mechanical oscillator. \\
MultiChua               &  Mufcstak E. Yalcin, Johan A. K. Suykens, Joos ... &     Multiple interacting Chua electronic circuits. \\
NewtonLiepnik           &  Leipnik, R. B., and T. A. Newton (1981). Doubl... &  Euler's equations for a rigid body, augmented ... \\
NoseHoover              &  Nose, S (1985). A unified formulation of the c... &  Fixed temperature molecular dynamics for a str... \\
NuclearQuadrupole       &  Baran V. and Raduta A. A. (1998), Internationa... &  A quadrupole boson Hamiltonian that produces c... \\
OscillatingFlow         &  T. H. Solomon and J. P. Gollub, Phys. Rev. A 3... &  A model fluid flow that produces KAM tori. Ori... \\
PanXuZhou               &  Zhou, Wuneng, et al. On dynamics analysis of a... &  A named attractor related to the DequanLi attr... \\
PehlivanWei             &  Pehlivan, Ihsan, and Wei Zhouchao (2012). Anal... &  A system with quadratic nonlinearity, which un... \\
PiecewiseCircuit        &  A. Tamasevicius, G. Mykolaitis, S. Bumeliene, ... &  A delay model that can be implemented as an el... \\
Qi                      &  G. Qi, M. A. van Wyk, B. J. van Wyk, and G. Ch... &  A hyperchaotic system with a wide power spectrum. \\
QiChen                  &         Qi et al. Chaos, Solitons \& Fractals 2008. &  A double-wing chaotic attractor that arises fr... \\
RabinovichFabrikant     &  Rabinovich, Mikhail I.; Fabrikant, A. L. (1979... &  A reduced-order model of propagating waves in ... \\
RayleighBenard            & Yanagita, Kaneko (1995). Rayleigh-Bénard...  &        A reduced-order model of a convective cell. \\
RikitakeDynamo          &  Rikitake, T., Oscillations of a system of disk... &  Electric current and magnetic field of two cou... \\
Rossler                 &  Rossler, O. E. (1976), An Equation for Continu... &    Spiral-type chaos in a simple oscillator model. \\
Rucklidge               &  Rucklidge, A.M. (1992). Chaos in models of dou... &  Two-dimensional convection in a horizontal lay... \\
Sakarya                 &  Li, Chunbiao, et al (2015). A novel four-wing ... &  An attractor that arises due to merging of two... \\
SaltonSea               &  Upadhyay, Bairagi, Kundu, Chattopadhyay (2007)... &  An eco-epidemiological model of bird and fish ... \\
SanUmSrisuchinwong      &                San-Um, Srisuchinwong. J. Comp 2012 &  A two-scroll attractor arising from dynamical ... \\
ScrollDelay             &  R.D. Driver, Ordinary and Delay Differential E... &  A delay model that can be implemented as an el... \\
ShimizuMorioka          &       Shimizu, Morioka. Phys Lett A. 1980: 201-204 &  A system that bifurcates from a symmetric limi... \\
SprottA                 &  Sprott (1994). Some simple chaotic flows. Phys... &  A member of the Sprott family of algebraically... \\
SprottB                 &  Sprott (1994). Some simple chaotic flows. Phys... &  A member of the Sprott family of algebraically... \\
SprottC                 &  Sprott (1994). Some simple chaotic flows. Phys... &  A member of the Sprott family of algebraically... \\
SprottD                 &  Sprott (1994). Some simple chaotic flows. Phys... &  A member of the Sprott family of algebraically... \\
SprottDelay             &  Sprott, J. C (2007). A simple chaotic delay di... &  An algebraically simple delay equation. A stan... \\
SprottE                 &  Sprott (1994). Some simple chaotic flows. Phys... &  A member of the Sprott family of algebraically... \\
SprottF                 &  Sprott (1994). Some simple chaotic flows. Phys... &  A member of the Sprott family of algebraically... \\
SprottG                 &  Sprott (1994). Some simple chaotic flows. Phys... &  A member of the Sprott family of algebraically... \\
SprottH                 &  Sprott (1994). Some simple chaotic flows. Phys... &  A member of the Sprott family of algebraically... \\
SprottI                 &  Sprott (1994). Some simple chaotic flows. Phys... &  A member of the Sprott family of algebraically... \\
SprottJ                 &  Sprott (1994). Some simple chaotic flows. Phys... &  A member of the Sprott family of algebraically... \\
SprottJerk              &  Sprott, J. C. Simplest dissipative chaotic flo... &  An algebraidally simple flow depending on a th... \\
SprottK                 &  Sprott (1994). Some simple chaotic flows. Phys... &  A member of the Sprott family of algebraically... \\
SprottL                 &  Sprott (1994). Some simple chaotic flows. Phys... &  A member of the Sprott family of algebraically... \\
SprottM                 &  Sprott (1994). Some simple chaotic flows. Phys... &  A member of the Sprott family of algebraically... \\
SprottMore              &  Sprott, J. C. (2020). Do We Need More Chaos Ex... &   A multifractal system with a nearly 3D attractor \\
SprottN                 &  Sprott (1994). Some simple chaotic flows. Phys... &  A member of the Sprott family of algebraically... \\
SprottO                 &  Sprott (1994). Some simple chaotic flows. Phys... &  A member of the Sprott family of algebraically... \\
SprottP                 &  Sprott (1994). Some simple chaotic flows. Phys... &  A member of the Sprott family of algebraically... \\
SprottQ                 &  Sprott (1994). Some simple chaotic flows. Phys... &  A member of the Sprott family of algebraically... \\
SprottR                 &  Sprott (1994). Some simple chaotic flows. Phys... &  A member of the Sprott family of algebraically... \\
SprottS                 &  Sprott (1994). Some simple chaotic flows. Phys... &  A member of the Sprott family of algebraically... \\
SprottTorus             &                      Sprott Physics Letters A 2014 &  A multiattractor system that goes to a torus o... \\
StickSlipOscillator     &  Awrejcewicz, Jan, and M. M. Holicke (1999). In... &  A weakly forced (quasiautonomous) oscillator w... \\
SwingingAtwood          &  Tufillaro, Nicholas B.; Abbott, Tyler A.; Grif... &  A mechanical system consisting of two swinging... \\
Thomas                  &  Thomas, Rene (1999). Deterministic chaos seen ... &  A cyclically-symmetric attractor correspondng ... \\
ThomasLabyrinth         &  Thomas, Rene. Deterministic chaos seen in term... &  A system in which trajectories seemingly under... \\
Torus                   &  See, for example, Strogatz (1994). Nonlinear D... &  A minimal quasiperiodic flow on a torus. All l... \\
Tsucs2                  &  Pan, Zhou, Li (2013). Synchronization of Three... &  A named attractor related to the DequanLi attr... \\
TurchinHanski           &  Turchin, Hanski. The American Naturalist 1997.... &   A chaotic three species food web. The species... \\
VallisElNino            &  Vallis GK. Conceptual models of El Nio and the... &  Atmospheric temperature fluctuations with annu... \\
VossDelay               &  Voss (2002). Real-time anticipation of chaotic... &  An electronic circuit with delayed feedback. A... \\
WangSun                 &  Wang, Z., Sun, Y., van Wyk, B. J., Qi, G. \& va... &                            A four-scroll attractor \\
WindmiReduced           &  Smith, Thiffeault, Horton. J Geophys Res. 2000... &  Energy transfer into the ionosphere and magnet... \\
YuWang                  &  Yu, Wang (2012). A novel three dimension auton... &  A temperature-compensation circuit with an ope... \\
YuWang2                 &  Yu, Wang (2012). A novel three dimension auton... &  An alternative temperature-compensation circui... \\
ZhouChen                &  Zhou, Chen (2004). A simple smooth chaotic sys... &                          A feedback circuit model. \\
\toprule
\end{longtable}
\clearpage

\section{Dataset structure and format}

All systems are primarily represented as Python objects, with names matching those in Figure \ref{fig_all} and the accompanying table. Underlying mathematical properties, parameters of the governing differential equation, recommended integration timestep and period, and default initial conditions are accessed as instance attributes. A callable implementation of the right hand side of the differential equation, a function for loading precomputed trajectories, and a function for re-integrating with default initial conditions and timescales, are included as instance methods. Additionally, we include a separate submodule for loading precomputed time series in bulk, or re-integrating all systems, which are useful for benchmarking tasks.

Our object representation abstracts the underlying records and metadata for each system, which are stored in a \texttt{JSON} file. The attributes recorded in the database file for each system are listed in Table \ref{metadata}. 

For each dynamical system, we include $16$ precomputed time series corresponding to all combinations of the following: coarse and fine sampling granularity, train and test splits emanating from different initial conditions, multivariate and univariate views, and trajectories with and without Brownian noise influencing the dynamics. The precomputed granularities correspond to a coarse granularity sampled at $15$ points per period (the dominant timescale determined by surrogate testing on the power spectrum), and a fine granularity sampled at $100$ points per period. The stochastically-forced trajectories correspond to adding a Langevin forcing term to the right hand side of each term in the dynamical equation. We used a scaled force with amplitude equal to to $1/40$ the standard deviation of the values the dynamical variable takes on the attractor in the absence of noise. When integrating these trajectories, we use variant of the Runge-Kutta algorithm for stochastic differential equations \cite{rossler2010runge}, as implemented in the Python package \texttt{sdeint}.

\begin{table}
\caption{Properties recorded for each chaotic system in the dataset}
\hspace*{-2cm}
\begin{tabular}{ l l }
\hline
System Name & \null \\
\hline
Reference & A citation to published work or original source where available. \\
Description & A brief description of domain area, or original motivation for publication \\
Parameters & Parameters governing the differential equation (e.g for bifurcations) \\
Embedding Dimension & The number of dynamical variables, or the number set by default for delay equations \\
Unbounded Indices & Indices of dynamical variables that grow without bound (e.g. time for nonautonomous systems) \\
dt & The integration timestep, determined by surrogate testing of the power spectrum \\
Initial Conditions & Initial conditions on the attractor, determined by a long simulation discarding a transient \\
Period & The dominant timescale in the system, determined by surrogate testing of the power spectrum \\
Lyapunov Spectrum & The spectrum of Lyapunov exponents, measure of trajectory dispersion \\
Largest Lyapunov Exponent & The largest Lyapunov exponent, a measure of chaoticity \\
Correlation Dimension & The fractal dimension, a measure of geometric complexity \\
Kaplan-Yorke Dimension & An alternative fractal dimension, a measure of geometric complexity \\
Multiscale Entropy & A measure of signal complexity \\
Pesin Entropy & An upper bound on the entropy under discretized measurements \\
Delay & Whether the system is a delay differential equation \\
Hamiltonian & Whether the dynamics are Hamiltonian \\
Non-autonomous & Whether the dynamics depend explicitly on time\\
\hline
\end{tabular}
\label{metadata}
\end{table}

\section{Glossary}

Here, we provide a glossary of several terms as they appear in the work presented here. More detailed treatments can be found in several references \cite{guckenheimer2013nonlinear,kuznetsov2013elements,strogatz2018nonlinear,ott2002chaos}.

\paragraph{Attractor.} A set of points within the state space of a dynamical system that most initial conditions approach over time. These points usually represent a subset of the full state space. In the work presented here, “attractor” and “dynamical attractor” are used interchangeably. 

\paragraph{Bifurcation.} A qualitative change in the dynamics exhibited by a dynamical system, as one or more system parameters is varied. For example, strange attractor can become a periodic orbit or fixed point as one of the parameters of the underlying dynamical equations is varied. Importantly, bifurcations occur as the result of changes to the underlying dynamical system, and do not in themselves result from the dynamics.

\paragraph{Dynamical System.} A set of rules describing how points within a space evolve over time. Dynamical systems usually appear either as (1) systems of coupled ordinary differential equations, which can be integrated to produce continuous-time trajectories, or (2) discrete-time maps that send points at one timepoint to new points a fixed interval $\Delta t$ later. In the context of the work presented here, a dynamical system is a single set of deterministic ordinary differential equations (e.g. the Lorenz system). 

\paragraph{Entropy.} A statistical property of a dynamical system corresponding to the gain of information over time as the system is observed. A highly regular and predictable process will have low entropy, while a stochastic process will have high entropy. Unlike dimensionality, the entropy of a system typically does not require a notion of distance on the state space. For example, if different regions of an attractor are colored with discrete labels, it is possible to define the entropy of a trajectory based on the sequence of symbols it passes through—without referencing the precise locations visited, or the distance among the symbols.

\paragraph{Ergodic.} A property of a dynamical system specifying that, over sufficiently long timescales, the system will visit all parts of its state space. A dissipative dynamical system will not be ergodic over its full state space, but it may be ergodic once it settles onto an attractor. In the context of time series analysis, ergodicity implies that a forecasting model trained on many short trajectories initialized at different points on an attractor will have the same properties as a model trained on subsections of a single long trajectory.

\paragraph{Fractal.} A set of points that appears self-similar over all length scales. Fractals have dimensionality intermediate to traditional mathematical objects like lines and surfaces, resulting in a diffuse appearance.

\paragraph{Initial Conditions.} A point within the state space of a dynamical system. As time passes, the rules specifying the dynamical system will transmit this point to other points within the system’s state space. An initial condition does not necessarily lie on an attractor of the dynamical system.

\paragraph{Limit Cycle.} A type of attractor in which trajectories undergo recurring periodic motion. A swinging, frictionless pendulum exhibits a limit cycle.

\paragraph{Lyapunov Exponent.} The initial growth rate of an infinitesimal perturbation to a point within a dynamical system’s state space. If two initial conditions are chosen with infinitesimal initial separation, then as time passes the two points will spread apart exponentially. The logarithm of the rate of change in their separation equals the Lyapunov exponent. For non-chaotic systems (such as systems evolving along regular limit cycles), neighboring points do not diverge, and so the Lyapunov exponent is zero. When used in reference to an entire attractor, the Lyapunov exponent corresponds to an average over all points on the attractor.

\paragraph{Quasiperiodic Motion.} A type of attractor corresponding to non-repeating continuous motion, which does not exhibit fractal structure. The dynamics contain at least two frequencies that are incommensurate with one another. Quasiperiodic attractors have integer fractal dimension and a surface-like appearance, in contrast to the diffuse appearance of strange attractors.

\paragraph{Stable Fixed Point.} A type of attractor in which trajectories converge to a single location within the state space. 

\paragraph{State Space.} The set of all possible states of a dynamical system. Initial conditions, trajectories, and attractors are all subsets of this space.

\paragraph{Strange Attractor.} An attractor in which trajectories continuously wander over a bounded region in state space, but never stop at a fixed point or settle into a repeating limit cycle. The dynamics are therefore globally stable, but locally unstable: the attractor contains a dense set of unstable periodic orbits, and trajectories briefly shadow individual orbits before escaping onto others. These unstable orbits span a continuous range of frequencies, producing motion at a range of length scales—and resulting in the fractal appearance of strange attractors.

\paragraph{Trajectory.} A set of points corresponding to the locations to which a given initial condition is mapped by a dynamical system. Trajectories are continuous curves for continuous-time systems, and isolated points for discrete-time maps.

\section{Calculation of mathematical properties}

For all mathematical properties we perform $20$ replicate computations from different initial conditions, and record the average in our database. To ensure high-quality estimates, we compute trajectories at high granularity of $500$ points per period (as determined by the dominant frequency in the power spectrum), and we use trajectories with length $2500$, corresponding to five complete periods.

\paragraph{Timescale alignment.} All systems in our database have been timescale-aligned, allowing them to be re-integrated at equivalent dominant timescales and sampling rates. This feature differentiates our approach from other time series collections, as well as previous applications of data-driven models to ordinary differential equations, and it allows easier comparison among systems. In order to align timescales, for each system we calculate the optimal integration timestep by computing the power spectrum, and then using random phase surrogates in order to identify the smallest and dominant significant significant frequencies \cite{kantz2004nonlinear}. The smallest frequency determines the integration timestep when re-integrating each system, while the highest amplitude peak in the power spectrum determines the dominant significant frequency, and thus the governing timescale. We use the dominant timescale to downsample integrated dynamics, ensuring consistency across systems. We record both fields in our database.

\paragraph{Lyapunov Exponents.} We implement  standard techniques for computing Lyapunov exponents \cite{wolf1985determining,holzfuss1991lyapunov,datseris2018dynamicalsystems}. Our basic approach consists of following a bundle of vectors along a trajectory, and at each timestep using the Gram-Schmidt procedure to re-orthonormalize the bundle. The stretching rates of the principal axes provide estimates of the Lyapunov exponents in each direction.

When determining the Lyapunov exponents, for each initial condition we continue integration until the smallest-magnitude Lyapunov exponent drops below our tolerance level of $10^{-8}$, because all continuous time systems have at least one zero-magnitude exponent. Our replicate spectrum estimates across initial conditions are averaged with weighting proportional to the distance between the smallest magnitude exponent and zero, in order to produce a final estimate.

\paragraph{Fractal Dimension.} We compute the fractal dimension using the Grassberger-Procaccia algorithm for the correlation dimension, a robust nonparametric estimator of the fractal dimension that can be calculated deterministically from finite point sets \cite{grassberger1983characterization}.

\paragraph{Entropy.} The multiscale entropy was used to estimate the intrinsic complexity of each trajectory \cite{costa2002multiscale}. While a multivariate generalization of the multiscale entropy has recently been proposed \cite{ahmed2011multivariate}, due to convergence issues we calculate the entropy separately for each dynamical variable, and then record the median across all coordinates. Because this approach fails to take into account common motifs across multiple dimensions, we expect that our calculations overestimate the true entropy of the underlying systems. A similar effect occurs when mutual information is computed among subsets of correlated variables.

\paragraph{Additional mathematical properties.} We derive and record in our database several properties derived from the spectrum of Lyapunov exponents, including the Pesin's upper bound on the entropy (the sum of all positive Lyapunov exponents) and the Kaplan-Yorke fractal dimension (an alternative estimator of the fractal dimension) \cite{ott2002chaos,kantz2004nonlinear}.

\section{Statistical Features and Embedding}

For each dynamical system, we generate $40$ trajectories of length $2000$ originating from random initial conditions on the attractor. We use the default granularity of $100$ points per dominant period as determined by Fourier transform. For each system and replicate, we compute $787$ standard common time series features using standard methods \cite{christ2018time}. For each dynamical system and replicate, we drop all null features, and then use an inner join operation to retain only features that appear across all dynamical systems and replicates. We then retain only the $100$ features with the highest variance relative to their mean values across all dynamical systems. 

We use these features to generate an embedding with UMAP \cite{mcinnes2018umap}. We repeat this procedure for each of the $40$ random initial conditions that were featurized for each dynamical system, and we report the median across replicates as the embedding of the dynamical system. We use affinity propagation with default hyperparameters in order to identify eight clusters within the embedding \cite{pedregosa2011scikit}.

\section{Forecasting Experiments}

Benchmarks are computed on the Harvard FAS Cannon cluster, using two Tesla V100-PCIE-32GB GPU and 32 GB RAM per node. Benchmarks are implemented with the aid of the \texttt{darts}, \texttt{GluonTS}, and \texttt{sktime} libraries \cite{loning2019sktime,alexandrov2020gluonts,unit8co2020darts}. 

\paragraph{Models.} We include forecasting models from several domains: deep learning methods (NBEATS, Transformer, LSTM, and Temporal Convolutional Network), statistical methods (Prophet, Exponential Smoothing, Theta, 4Theta), common machine learning techniques (Random Forest), classical forecasting methods (ARIMA, AutoARIMA, Fourier transform regression), and standard naive baselines (naive mean, naive seasonal, naive drift) \cite{oreshkin2019n,lea2016temporal,alexandrov2020gluonts,godahewa2021monash}. All non-tuned hyperparameters (e.g. training epochs, number of layers, etc) are kept at default values used in reference implementations included in the \texttt{darts}, \texttt{GluonTS}, and \texttt{sktime} libraries \cite{loning2019sktime,alexandrov2020gluonts,unit8co2020darts}.

\paragraph{Hyperparameter tuning.} Hyperparameter tuning is performed separately for each forecasting model, dynamical system, and sampling granularity. The training set for each attractor consists of a single train time series comprising a trajectory emanating from a random location on the chaotic attractor. For each trajectory, $10$ full periods are used to train the model, and $2$ periods are used to generate forecast mean-squared-errors to evaluate combinations of hyperparameters. These splits correspond to $150$ and $30$ timepoints for the coarse granularity datasets, and $1000$ and $200$ timepoints for the fine granularity datasets.

Because benchmarks are computed on both coarse and fine granularities, different value ranges are searched for the two granularities: 1 timepoint, 5 timepoints, half of a period (8 timepoints for the coarse granularity, 50 timepoints for the fine granularity), and one full period (15 timepoints / 100 timepoints). For forecast models that accept a seasonality hyperparameter, the presence of additive seasonality (such as monochromatic forcing) is treated as an additional hyperparameter. A standard grid search is used to find the best sets of hyperparameters separately for each model, system, and granularity.

\paragraph{Scoring.} The testing dataset consists of a single time series emanating from another point on the same attractor. On this trajectory, a model is trained on the first $10$ periods using the best hyperparameters the train dataset, and the forecast score is generated on the remaining $2$ periods of the testing time series. Several standard time series similarity metrics are recorded for each dynamical system and forecasting model: mean absolute percentage error (MAPE), symmetric mean absolute percentage error SMAPE, coefficient of variation (CV), mean absolute error (MAE), mean absolute ranged relative error (MARRE), mean squared error (MSE), root mean squared error (RMSE), coefficient of determination ($r^2$), and mean absolute scaled error (MASE).

\subsection{The effect of noise on forecasting results.} 

\begin{figure*}
  \centering
 \includegraphics[width=\linewidth]{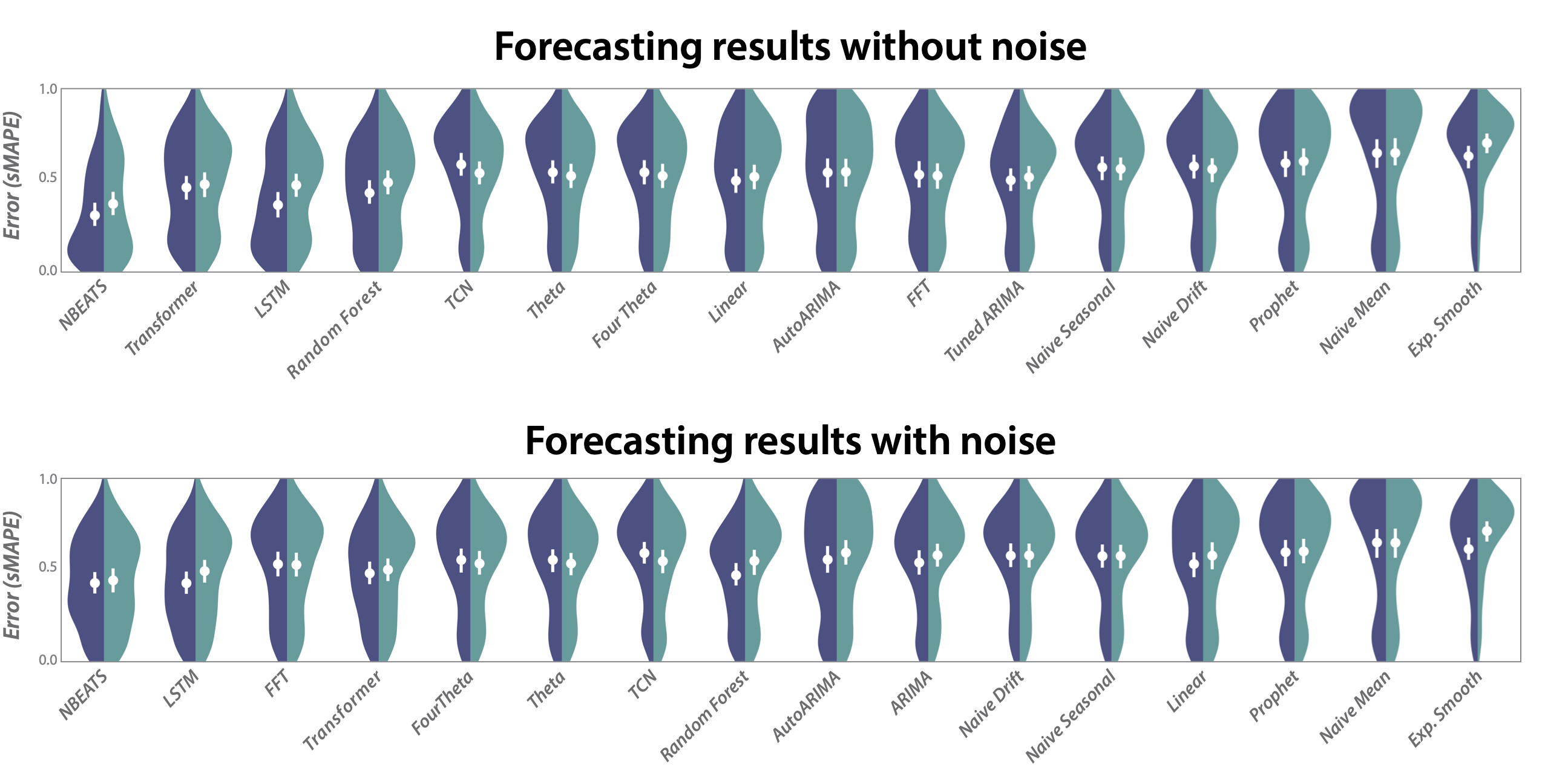}
  \caption{{\bf Forecasting results with and without noise.} Each panel shows the distribution of forecast errors for all dynamical systems across different forecasting models, sorted by increasing median error. Dark and light hues correspond to coarse and fine time series sampling granularities. Upper panel corresponds to results for the full chaotic systems collection without noise, and lower panel corresponds to results from replicate experiments in which noise is present. Note that the model order along the horizontal axis differs between the two panels, because the relative performance of the different forecasting methods changes in the presence of noise.
 }
 \label{fig_noise_forecast}
\end{figure*}

In order to determine the robustness of our experimental results to the presence of non-deterministic noise in the dataset, we perform a full replication of our experiments above on a modified dataset that includes noise. For each dynamical system, the scale of each dynamical variable is determined by generating a reference trajectory without noise, and calculating the standard deviation along each dimension. A new trajectory is then generated with noise of amplitude equal to 20\% of the scale of each dynamical variable. Figure \ref{fig_noise_forecast} shows the result of our benchmarks with noise, compared to our benchmarks in the absence of noise.

As expected, the median forecasting performance degrades for all methods in the presence of noise. Noise only weakly affects the naive baselines, because the range of values present in the data remains the same in the presence of noise. The deep learning models continue to perform very well, consistent with general intuition that large, overparametrized models effectively filter low-information content from complex signals \cite{hastie2009elements}. Interestingly, the performance of the random forest model noticeably degrades with noise, suggesting that the representation learned by the model is fragile in the presence of extraneous information from noise. Conversely, the simple Fourier transform regression performs better than several more sophisticated models in the presence of noise. We hypothesize that high-frequency noise disproportionately obfuscates phase information within the signal, and so forecasting models that project time series onto periodic basis functions (e.g., Fourier and N-BEATS) are least impacted.

\section{Forecasting experiments as granularity and noise are varied}

In order to better understand how the performance of different forecasting models depends on properties of the time series, we perform a set of experiments in which we re-train all forecasting models on datasets with a range of granularities and noise levels. We define noise level the same way as in our forecasting experiments: a noise level of $0.2$ corresponds to a noise amplitude equal to $20\%$ of the normal standard deviation of the signal. Granularity refers to the number of points sampled per period, as defined by the dominant significant frequency in the power spectrum. For these experiments, the same hyperparameters are used as for the original forecasting experiments. However, for the granularity sweep, hyperparameters that have units equivalent to timescale (e.g. number of time lags, or input chunk size) are rescaled by the granularity.

\begin{figure*}
  \centering
 \includegraphics[width=0.7\linewidth]{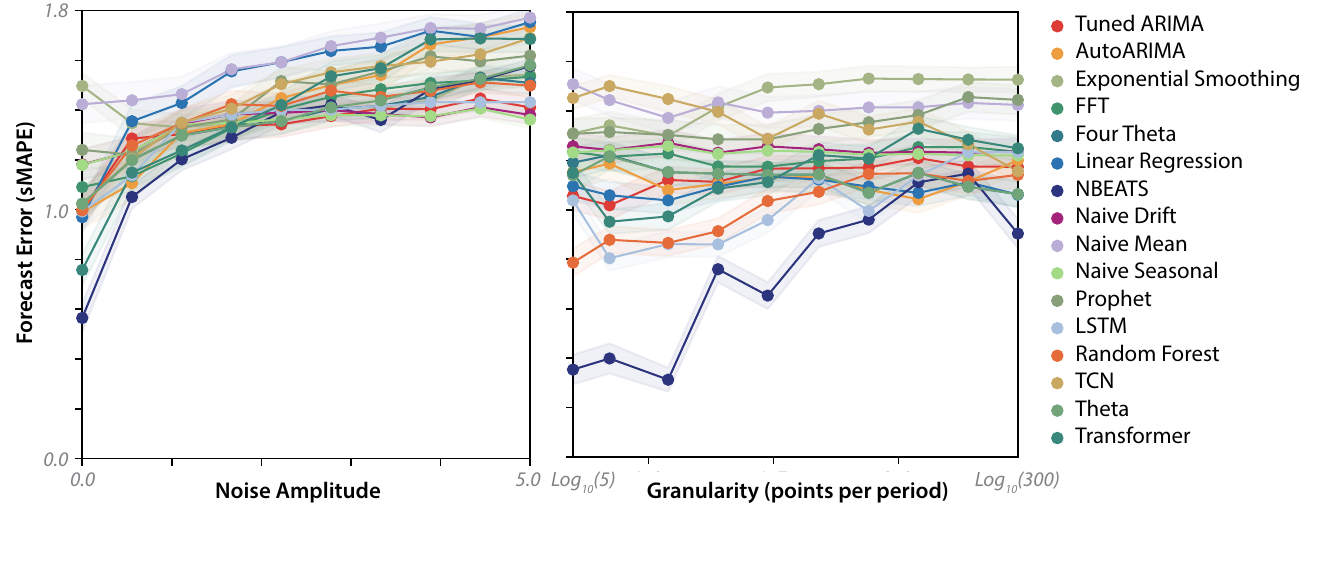}
  \caption{{\bf Variation in forecasting model performance as noise level and granularity are varied.} Points and shaded ranges correspond to medians and standard errors across dynamical systems.
 }
 \label{fig_sweep}
\end{figure*}

The results are shown in Figure \ref{fig_sweep}. We find that forecasting models are most strongly differentiated at low noise levels, and that as the noise level exceeds the average amplitude of the signal the performance of models converges. This effect arises because there is less useable information in the signal for forecasting. However, the relative ranking of the different models remains somewhat stable as noise intensity increases, suggesting that the deep learning models remain effective at extracting relevant information even in the presence of dominant noise.

The granularity results show that the relative performance of different forecasting models is stable across granularities, and that the deep learning models (and particularly NBEATS) continue to perform well across a range of granularities. However, unlike the statistical methods, the performance of the deep learning models fluctuates widely across granularities, and in a systematic manner that cannot be attributed to sampling error---all points and rankings are averages over all $131$ systems. These results suggest that more complex models may have timescale bias in their default architectures. However, we caution that exhaustive (albeit computationally expensive) hyperparameter tuning is needed to further understand this effect.

\section{Relative performance of forecasting models across different mathematical properties}

\begin{figure*}
  \centering
 \includegraphics[width=\linewidth]{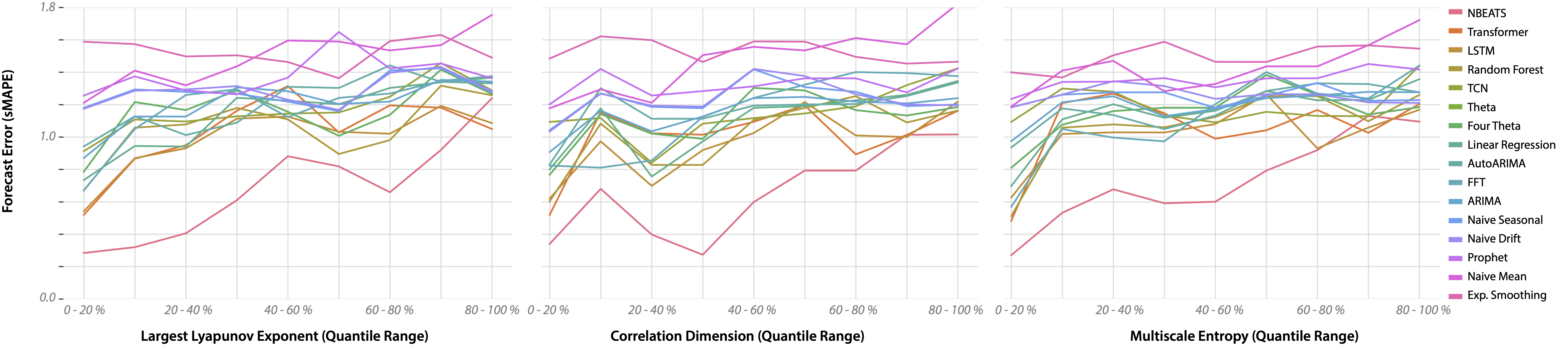}
  \caption{{\bf Variation in forecasting model performance across different mathematical properties.} The horizontal axis of each plot corresponds to a sliding window comprising a $20\%$ quantile in the property across all systems. Points correspond to medians across all dynamical systems in that quantile.
 }
 \label{fig_rank}
\end{figure*}

In order to determine whether different forecasting models are better suited to different types of dynamical system, we analyze our forecasting benchmarks striated by different mathematical properties of the dynamical systems. For a given mathematical property (such as Lyapunov exponent), we select only the dynamical systems among the bottom $20\%$ of systems (i.e. the least chaotic systems), and we compute the average forecast error for each forecasting model on just this group. We repeat the analysis for the dynamical systems in the quantile $10 - 30\%$, then $20 - 40\%$, and so forth in order to determine how forecasting performance of each model type varies with level of chaoticity. We repeat the analysis for the correlation dimension and multiscale entropy. Our results are shown in Figure \ref{fig_rank}

\section{Importance Sampling Experiments}

Our importance sampling experiment consists of a modified version of our forecasting task. We choose a single model, the LSTM, and alter its training procedure in order to determine how it is affected by alternative sampling strategies. In order to control for unintended interactions, we use a single set of hyperparameters for models trained on all chaotic systems, corresponding to the most common values from our forecasting benchmark. As a result, the baseline forecast error is higher across the chaotic systems dataset compared to our forecasting experiments, in which the LSTM was tuned separately for each chaotic system.

Our procedure consists of the following: (1) We halt training every few epochs and compute historical forecasts (backtests) on the training trajectory. (2) We randomly sample timepoints proportionately to their error in the historical forecast, and then generate a set of initial conditions corresponding to random perturbations away from each sampled attractor point. (3) We simulate the full dynamical system for $\tau = 150$ timesteps for each of these initial conditions, and we use these new trajectories as the training set for the next $b = 30$ epochs. We repeat this procedure for $\nu = 5$ meta-epochs. For the original training procedure, the training time scales as $\sim B = 400$, the number of training epochs times the number of timepoints in a full trajectory. 

For the control "full epoch" baseline, we use the standard training procedure. For our "random batch" control experiments, we repeat the importance sampling procedure, but randomly sample timepoints, rather than weighting points by their backtest error. We include this control in order to account for the possibility of forecast error decreasing with total training data, an effect that would lead the importance sampling procedure to perform well spuriously.

\section{Transfer Learning Experiments}

For our classification experiments, we start with the $128$ tasks currently within the UCR time series classification archive, and we narrow the set to the $96$ datasets that contain at least $100$ valid timepoints \cite{dau2019ucr}.

Our autoencoder is based on a causal dilated architecture recently shown to provide competitive performance among unsupervised embedding methods on the UCR archive \cite{franceschi2019unsupervised}. Following previous work, our encoder comprises a single causal convolutional block \cite{bai2018empirical}, containing two causal convolutions with kernel size $3$ and dilations of $2$. A convolutional residual connection bridges the input layer and the latent layer, and leaky ReLU activations are used throughout. Unlike previous studies that learned embeddings using a triplet loss (thereby eliminating the need for a decoder) \cite{franceschi2019unsupervised}, we use a standard decoder similar to our previous study on chaotic system embedding \cite{gilpin2020deep}, consisting of a three-layer standard convolutional network with ELU activation functions. We train our models using the Adam optimizer with mean squared error loss and a learning rate of $10^{-3}$ \cite{kingma2014adam}. Our PyTorch network implementations are included in the project repository.

We train separate encoders for each classification task in the UCR archive. Briefly, we retrieve the training dataset for a given classifation task, and we use phase surrogate testing to determine the dominant frequency in the training data. We then convert this timescale into an effective granularity (in points per dominant period) for the training data. We then re-integrate all $131$ dynamical systems within our dataset, with a granularity setting set to match the training data. We train the autoencoder on these trajectories, and we then apply the encoder to the training data of the classification task, in order to generate a featurized time series. For our "random timescale" ablation experiment, we select random granularities unrelated to the training data, and otherwise repeat the procedure above.

Having obtained encoded representations of the classification task training data, we then convert the training data into a featurized representation using \texttt{tsfresh}, a suite that generates $787$ standard time series features (such as number of peaks, average power, wavelet coefficients) \cite{christ2018time}. We then pass these features to a standard ridge regression classifier, which we set to search for $\alpha$ values over a range $10^{-3}$ -- $10^{3}$ via cross-validation \cite{pedregosa2011scikit}. Our approach to classifying time series is based upon recent methods for generating classification results from features learned from time series in an unsupervised setting, which found that complex unsupervised feature extractors followed by supervised linear classification yield competitive performance \cite{dempster2020rocket}. For our "no transfer learning" baseline, we apply the featurization and regression to the bare original training data for the classification problem.

Our reported scores correspond to accuracy on the test partition of the UCR archive. The timescale extraction, surrogate data generation, autoencoder, \texttt{tsfresh} featurization, and ridge classifier cross-validation steps are all trained only on the training data, and the trained encoder, \texttt{tsfresh} featurization, and ridge classifer are applied to the test data.

\section{Symbolic Regression Experiments}

Our symbolic regression dataset consists of input values corresponding to points along a trajectory, and target values corresponding to the value of the right hand side of the governing differential equation at those points. For our benchmark, we generate train and test datasets corresponding to trajectories originating from different locations on the attractor. Because we are interested in performance using information sampled across the attractor, we generate long trajectories ($10$ full periods, as determined by dominant timescale in power spectrum) at low sampling granularity ($15$ points per period), for a total of $150$ datapoints in each of the train and test trajectories. This number of points is comparable to existing benchmarks \cite{la2021contemporary}. While, in principle, random inputs could be generated and used to produce output values for our differential equations, because our target formulae correspond to dynamical systems, we favor using trajectories---which would best simulate observations from a real-world system. As we note in the main text, the accuracy of the target formulae will likely be reduced in regions of the attractor with lower measure.

For PySINDY, we fit separate models with purely polynomial and purely trigonometric bases. For DSR and pySR, we use default hyperparameters, and allow a fixed library of binary and unary expressions, $\{+, - , \times, \divsymb\}$, $\{\sin, \cos, \exp, \log, \tanh \}$ \cite{petersen2019deep}. Because our dynamical systems are multivariate, we fit separate expressions to each dynamical variable, and record the median across dynamical variables as the overall error for the system. 

We apply the expressions generated by symbolic regression to the unseen test trajectory, and we treat the resulting values as forecasts. We therefore record the same error metrics as for our forecasting benchmark above.

\section{Neural Ordinary Differential Equation Experiments}

We perform a preliminary neural ordinary differential equation (nODE) experiment, in order to evaluate whether mathematical properties of a dynamical system influence the properties of a fitted nODE. We design our experiment identically to our fine-granularity forecasting benchmark above: for each system, a multivariate training trajectory consisting of $1000$ timepoints is used to train a nODE model \cite{chen2018neural}. An unseen "test" initial condition is then randomly chosen, and $200$ timepoint trajectories are generated using both the true dynamical system, and the trained neural ODE. The quality of the resulting trajectory is evaluated using the sMAPE error between the predicted and true trajectory.

\begin{figure*}
  \centering
 \includegraphics[width=0.4\linewidth]{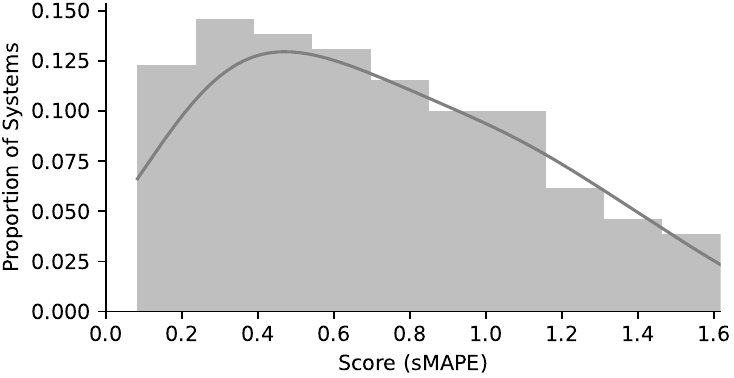}
  \caption{Distribution of error scores for the neural ordinary differential equations benchmark.
 }
 \label{fig_node}
\end{figure*}

Our results are shown in Figure \ref{fig_node}. Overall, the forecasting performance of the nODE model is competitive with other time series forecasting techniques, with the advantage of producing a differentiable representation of the underlying process that can potentially be used for downstream analysis. Qualitatively, we observe that the nODE dynamics frequently become trapped near unstable periodic orbits over long durations, suggesting that shadowing events observed in the training data dominate the learned representation \cite{guckenheimer2013nonlinear}.

Unlike our symbolic regression experiments, we find that there is no significant correlation between the quality of a nODE model and any underlying properties of the differential equations. Among the various mathematical properties (Lyapunov exponents, fractal dimension, etc) the largest observed Spearman correlation was not significantly different from zero ($0.072 \pm 0.003$, median with standard error determined by bootstrapping),

\section{Datasheet: Dataset documentation and intended uses}

The primary inclusion criteria for dynamical systems is appearance in published work with explicit equations and parameter values provided that created chaotic dynamics. While there are infinite possible chaotic attractors, our collection surveys systems as they appear in the literature---which primarily comprises particular domain-area applications, as well as systems with particular mathematical properties. Below, we address the questions included in an existing dataset datasheet guide \cite{gebru2018datasheets}.

\subsubsection{Motivation}
\label{motivation}

\textbf{Purpose} This dataset was created for the purpose of providing a generative benchmark for time series mining applications, in which arbitrary synthetic data can be generated using a deterministic process.

\textbf{Unintended Uses} To our knowledge, there are no pressing uses for this data that could cause unintended harm. However, insofar as our dataset can be used to improve existing time series models (illustrated by our time series classification benchmark), there is a possibility of our dataset contributing to privacy concerns with time series analysis---particularly by making it possible for large models to identify latent factors that could, for example, de-anonymize physiological recordings \cite{shi2011privacy}. In our project repository, we include instructions asking users who become aware of any unintended harms to submit an issue on GitHub.

\textbf{Previous Uses} Some time series analysis utilities and specific systems in this repository were used in our previous work \cite{gilpin2020deep}, but the full dataset and benchmarks are all new.

\textbf{Creator and Funding.} This repository was created by William Gilpin, with support from the NSF-Simons Center for Quantitative Biology at Harvard University, as well as the University of Texas at Austin. No special funding was solicited for this project.

\subsubsection{Composition}\label{composition}

\textbf{Instances.} Each instance in this dataset comprises a set of nonlinear differential equations describing a chaotic process, a set of standard parameter values and initial conditions, a set of default timescales and integration timesteps, a set of characteristic mathematica properties, a citation to a published source (where available), a brief description of the system, and $16$ precomputed trajectories from the system under various granularities and initial conditions.

\textbf{Instance Relationships.} Each instance corresponds to a different dynamical system.

\textbf{Instance Count.} At time of writing, there are 131 continuous-time dynamical systems (126 ordinary differential equations, and 5 delay equations). There are also 30 discrete-time chaotic maps, however we do not include these in any analyses or discussion presented here.

\textbf{Instance Scope.} Each instance corresponds to a particular realization of a dynamical system, based on previously-published parameter values and initial conditions. In principle, an infinite number of additional chaotic systems exists; our dataset seeks to provide a representative sample of published systems.

\textbf{Labels.} Each trajectory and system contains metadata describing its provenance, however there is not a particular label associated with each trajectory. However, all systems are labelled a variety of annotations that can, in principle, be used as labels (see Table \ref{metadata}).

\textbf{External Dependencies.} The data itself has no external dependencies. Simulating each system requires several standard scientific Python packages (enumerated in the repository README file). Running the benchmarks requires several additional dependencies, which are also listed in the README.

\textbf{Data Splits.} No splits are baked-in, because (in principle) arbitrary amounts of training, validation, and testing data can be generated for each dynamical system. Splits can either be performed by
holding out some timepoints, or (for multivariate systems) by splitting the set of dynamical variables. For the purpose of benchmarking experiments, splits corresponding to $10$ periods of training data, and
$2$ periods of unseen prediction/validation data, were used for both the train and test datasets (the test dataset corresponds to an unseen initial condition). For the fine granularity time series, this corresponds to splits of 1000/200 for both the train and test initial conditions. For the coarse granularity time series, this corresponds to a split of 150/30. The data loader utilities included in the Python library use the $10$ period / $2$ period split by default.

\textbf{Experiments.} All benchmark experiments are described at length
in our preprint. They primarily consist of forecasting benchmarks,
generative experiments (importance sampling and model pretraining), and
data-driven model inference experiments.

\subsubsection{Collection}\label{collection}

\textbf{Collection.} ISI Web of Science was used to identify papers claiming novel low-dimensional chaotic systems published after 1963 (the year of Lorenz's original paper). Papers were sorted by citations in order to determine priority for re-implementation, and systems were only included that had (1) explicit analytical expressions and (2) published parameter values and initial conditions leading to chaos. All systems were re-implemented in Python and checked to verify that the reported dynamics were chaotic. Additionally, several previous collections and galleries of chaos were checked, to ensure that all entries are included \cite{sprott2010elegant,maier2003,datseris2018dynamicalsystems,myers2020teaspoon}.

\textbf{Workers.} All individuals involved in data collection and
curation are authors on the paper.

\textbf{Timeframe.} Data was collected from 2018 -- 2021.

\textbf{Instance Acquisition.} Each dynamical system required
implementation in Python of the stated dynamical equations, as well as
all parameter values and initial conditions leading to chaos. Each
system was then numerically integrated in order to ensure that the
observed dynamics matched those claimed in the original publication.
Once chaos was validated, the integration timestep and the trajectory
sampling rate were determined using the power spectrum, with time series
surrogate analysis used to identify significant frequencies. Once the
correct timescales were known, properties such as the Lyapunov exponents
and entropy were calculated. For all trajectory data and initial
conditions, a long transient was discarded in order to ensure that the
dynamics settled onto the attractor.

\textbf{Instance Scope.} There are effectively an infinite number of
possible chaotic dynamical systems, even in low dimensions. However, our
collection represents a sample of named and published chaotic systems,
and it includes most well-known systems.

\textbf{Sampling.} Because our dataset comprises only named and
published chaotic systems, it does not comprise a representative sample
of the larger space of all low-dimensional chaotic systems. Therefore,
our database should not be used to compute any quantities that depend on
the measure of chaotic systems within the broader space of all possible
dynamical systems. For example, a study that seeks to identify the most
common features or motifs of chaotic systems cannot use our database as
representative sample. However, our database does comprise a
representative sample of chaotic dynamics as they appear in the
literature.

\textbf{Missing Information.} For systems in which a reference citation
or additional context is unavailable, the corresponding field in the
metadata file is left blank. However, all systems have sufficient
information to be integrated.

\textbf{Errors.} If any errors or redundancies are identified,
we encourage users to submit an issue via GitHub. 

\textbf{Noise.} Noise can be added to the trajectories either by adding random values to each observed
timepoint (measurement noise), or performing a stochastic simulation (stochastic dynamics). A stochastic integration function is included in the Python library. The precomputed trajectories associated with each system include trajectories with noise.

\subsubsection{Preprocessing}\label{preprocessing}

\textbf{Cleaning.} Dynamical systems may be numerically integrated with
arbitrary precision, and their dynamics can be recorded at arbitrarily
small intervals. In order to report all systems consistently, we use
time series phase surrogate testing to identify the highest significant
frequency in the power spectrum of each system's dynamics. We then set
the numerical integration timestep to be proportional to this timescale.
We then re-integrate, and use surrogates to identify the dominant
significant frequency in each system's dynamics. We use this timescale
to determine the sampling rate. This process ensures overall that all
systems exhibit dynamical variation over comparable timescales, and that
the integration timestep is sufficiently small to accurately resolve the
dynamics.

Having determined the appropriate integration timescales, we then determine the Lyapunov exponents, average period, and other ensemble-level properties of each dynamical system. We compute these quantities for replicate trajectories originating from different initial conditions on the attractor, and record the average.

For each fixed univariate time series dataset, the first ordinal component of the system's dynamics is included.

\textbf{Raw data.} New time series data can be generated as needed via
the \texttt{make\_trajectory()} method of each dynamical system.

\textbf{Preprocessing Software.} All analysis software is included in
the repository.

\textbf{Motivation.} To our knowledge, dataset processing is consistent
with the underlying motivation of the dataset.

\subsubsection{Distribution}\label{distribution}

\textbf{Distribution.} The dataset is distributed on GitHub.

\textbf{First Distribution.} A private fork may be distributed with the
paper for review in order to maintain anonymity for certain venues. The
updated repository will be distributed with the final paper.

\textbf{License.} We include an Apache 2.0 License in the project
repository.

\textbf{Fees.} None.

\subsubsection{Legal}\label{legal}

\textbf{People.} No individuals are included in this dataset.

\textbf{Protected Subjects.} No ethically-protected subjects are
included in this dataset.

\textbf{Institutional Approval.} No institutional approval is required
for this dataset

\textbf{Consent.} No individual data is included in this dataset.

\textbf{Harm.} No individual data is included in this dataset. However, the README file of the dataset repository includes instructions to submit an issue if an unintended harm is detected in the process of using this dataset.

\textbf{Disadvantages.} No individual data is included in this dataset.

\textbf{Privacy.} None of the data contains personal information.

\textbf{GDPR.} To our knowledge, this dataset complies with GDPR and
equivalent foreign standards.

\textbf{Sensitivity.} To our knowledge, this dataset contains no
sensitive or confidential information

\textbf{Inappropriate.} This dataset contains no inappropriate or offensive content.

\section{Author statement and hosting plan}

The authors bear all responsibility in case of rights violations. The data license has been included elsewhere in this appendix. The authors have full control of the data repository on GitHub, and will ensure its continued accessibility.

\clearpage
\bibliography{dysts_cites} 
\bibliographystyle{naturemag}